\documentclass[journal,12pt,onecolumn,draftclsnofoot,]{IEEEtran}
\usepackage[T1]{fontenc}

\usepackage{tikz}
\usepackage{etoolbox}
\usepackage{enumitem}
\newcommand*{\circled}[1]{\lower.7ex\hbox{\tikz\draw (0pt, 0pt)%
    circle (.5em) node {\makebox[1em][c]{\small #1}};}}
\robustify{\circled}
\usepackage{amsmath}
\usepackage{hyperref}
\usepackage{mathrsfs}
\usepackage{mathrsfs}
\usepackage{amsfonts}
\usepackage{graphicx,cite,epsfig,amssymb,amsmath}
\usepackage{color,xcolor}
\usepackage{bm}
\definecolor{red}{rgb}{1.00, 0.00, 0.00}  
\usepackage{pifont}
\usepackage{stmaryrd}
\usepackage{setspace}
\usepackage{subfigure}
\usepackage{cite}
\usepackage{array}
\usepackage{float}
\usepackage{multirow}
\usepackage{algorithm,algpseudocode}
\usepackage{epstopdf}
\usepackage{mathtools}
\usepackage{amssymb}
\usepackage{diagbox}
\usepackage{booktabs}

\usepackage{makecell}
\usepackage{xcolor}
\makeatletter

\newcommand{\Rmnum}[1]{\expandafter\@slowromancap\romannumeral #1@}
\newcommand{\Zstroke}{%
  \text{\ooalign{\hidewidth\raisebox{0.2ex}{--}\hidewidth\cr$Z$\cr}}%
}
\makeatother
\usepackage{pifont} 
\usepackage{epstopdf}
\usepackage{mathtools}
\usepackage{diagbox}
\usepackage{booktabs}
\usepackage{multirow}
\usepackage{booktabs}
\usepackage{makecell}
\usepackage{float} 
\usepackage{subfigure} 
\usepackage{colortbl}

\usepackage{setspace}
\usepackage{tikz}
\begin{document}
\normalsize
\title{ Automatic AI Model Selection for Wireless Systems: Online Learning  via Digital Twinning}

\author{Qiushuo Hou \IEEEmembership{Graduate Student Member, IEEE}, Matteo Zecchin \IEEEmembership{Member, IEEE}, Sangwoo Park \IEEEmembership{Member, IEEE}, Yunlong Cai \IEEEmembership{Senior Member, IEEE}, Guanding Yu \IEEEmembership{Senior Member, IEEE}, Kaushik Chowdhury \IEEEmembership{Fellow, IEEE}, and Osvaldo Simeone \IEEEmembership{Fellow, IEEE}
\thanks{The work of M. Zecchin and O. Simeone is supported by the European
 Union’s Horizon Europe project CENTRIC (101096379). The work of O. Simeone is also supported by an Open Fellowship of the EPSRC (EP/W024101/1), and by the EPSRC project
 (EP/X011852/1). K. Chowdhury is supported by the NSF CCRI RFDataFactory project CNS \#2120447.
 
Q. Hou, Y. Cai, and G. Yu are with the College of Information Science and Electronic Engineering, Zhejiang University, Hangzhou 310027, China (e-mail: \{qshou, ylcai, yuguanding\}@zju.edu.cn).

Matteo Zecchin, Sangwoo Park, and Osvaldo Simeone are with the King’s Communications, Learning \& Information Processing (KCLIP) lab within the Centre for Intelligent Information Processing Systems (CIIPS), Department of Engineering, King’s College London, London WC2R 2LS, U.K. (e-mail: \{matteo.1.zecchin, sangwoo.park, osvaldo.simeone\}@kcl.ac.uk). 

Kaushik Chowdhury is with the Institute for the Wireless Internet of Things, Northeastern University, Boston, MA, USA  (e-mail:  k.chowdhury@northeastern.edu).
}}
\maketitle
\vspace{-1.5cm}
\begin{abstract}
In modern wireless network architectures, such as O-RAN, artificial intelligence (AI)-based applications are deployed at intelligent controllers to carry out functionalities like scheduling or power control. The AI ``apps'' are selected  on the basis of contextual information such as network conditions, topology, traffic statistics, and design goals. The mapping between context and AI model parameters is ideally done in a zero-shot fashion via an automatic model selection (AMS) mapping that leverages only contextual information without requiring any current data. This paper introduces a general methodology for the online optimization of AMS mappings. Optimizing an AMS mapping is challenging, as it requires exposure to data collected from many different contexts. Therefore, if carried out online, this initial optimization phase  would be extremely time consuming. A possible solution is to leverage a digital twin of the physical system  to generate synthetic data from multiple simulated contexts. However, given that the simulator at the digital twin is imperfect, a direct use of simulated data for the optimization of the AMS mapping would yield poor performance when tested in the real system. This paper proposes a novel method for the online optimization of AMS mapping that corrects for the bias of the simulator by means of limited real data collected from the physical system. Experimental results for a graph neural network-based power control app demonstrate the significant advantages of the proposed approach.

\end{abstract}
\section{Introduction}
\subsection{Context, Motivation, and Overview}

\begin{figure*}[htp]
    \centering
    \subfigure[run-time phase]{
    \includegraphics[width=7cm]{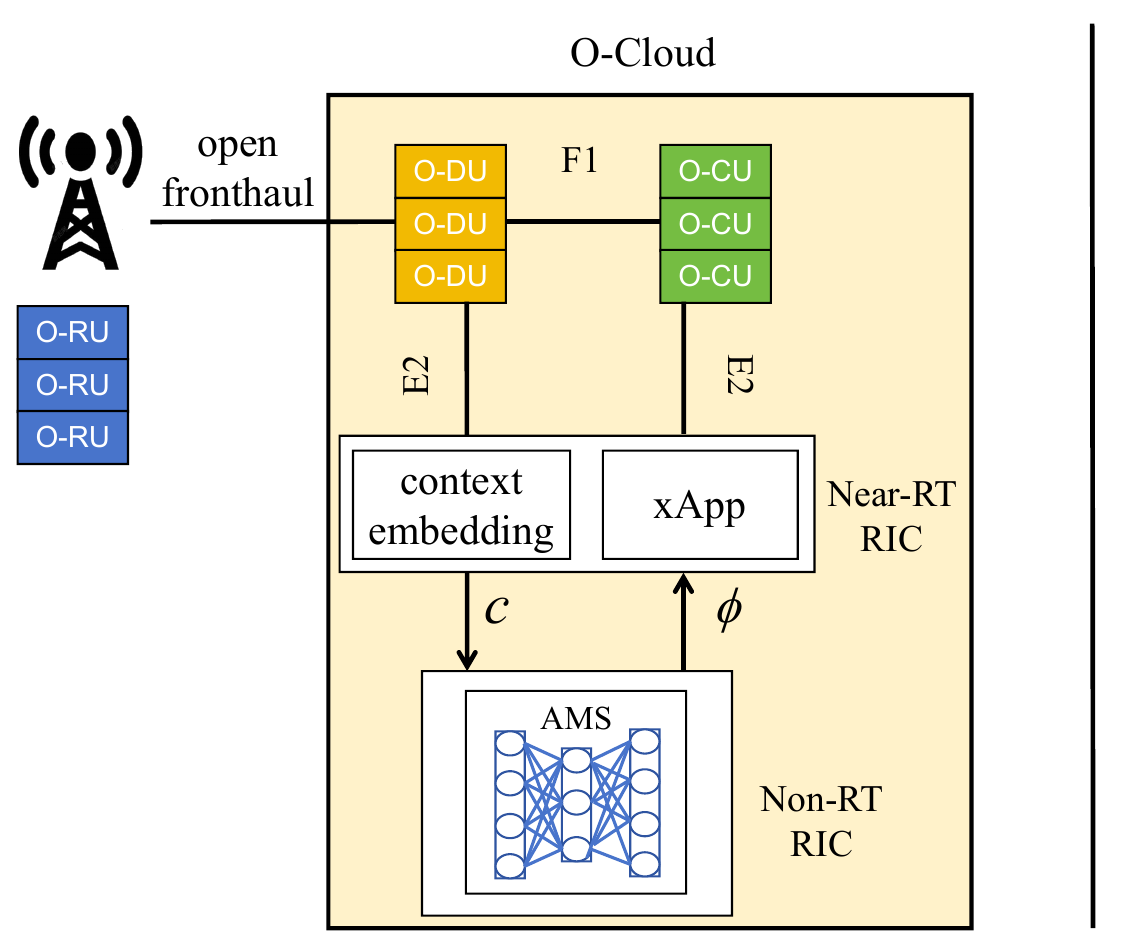}
    \vspace{-0.5cm}
    \label{fig: runtime}}
    \subfigure[calibration phase]
    {
    \centering
    \includegraphics[width=8cm]{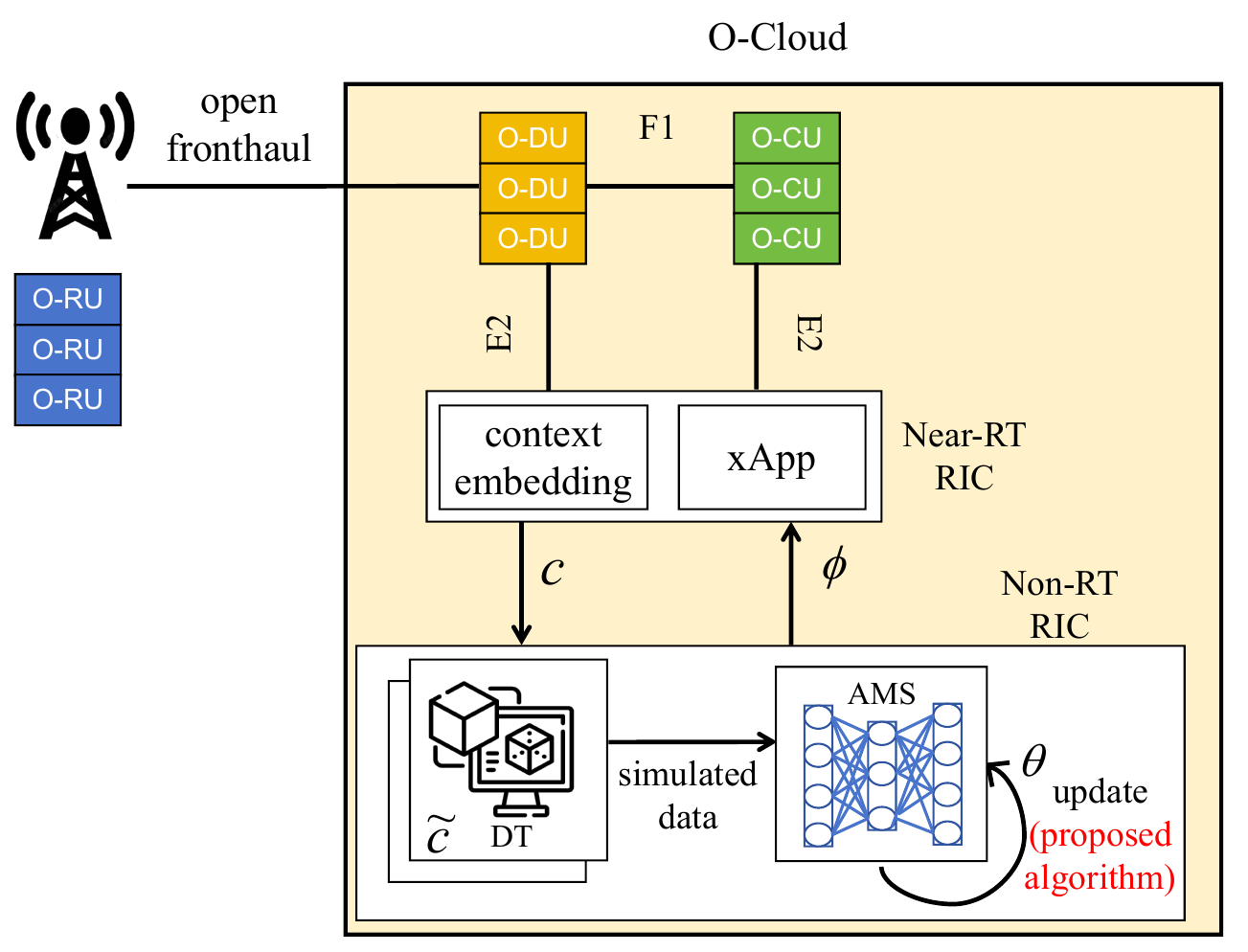}
    \label{fig: calibration}
    }
    \caption{As a use case of the proposed methodology, this figure illustrates the  O-RAN architecture, in which a base station (BS), known as gNB, is disaggregated into a central unit (CU), a distributed unit (DU), and a radio unit (RU), where both CU and DU are deployed in the O-Cloud, and the RU is deployed at the BS. The near-real time (RT) radio intelligent controller (RIC)  deploys AI-based xApps to carry out functionalities at different layers of the protocol stack. As shown in part (a), in the considered setting, an automatic model selection (AMS) mapping produces the parameters $\phi$ of an AI model  based on  context information from the BS. The AI model may be implemented at the near-RT 
 RIC for an O-RAN system. The focus of this work is on the online optimization of the AMS mapping, which is carried out during a preliminary calibration phase. Specifically, as shown in part (b) for the O-RAN architecture, this work proposes to speed up online calibration via the use of a digital twin, which can produce synthetic data for new contexts.  In the proposed scheme, for each real-world context $c$, the digital twin produces synthetic data from multiple simulated contexts, and the real-world data is leveraged to ``rectify'' the errors made by the simulator.}
    \label{fig: O-RAN}
\end{figure*}

In a modern wireless network, artificial intelligence (AI)-based applications are expected to carry out an increasingly large number of tasks at different layers of the protocol stack, such as scheduling, power control, and handover\cite{GNN1,GNN2,GNN3,handover}. For example, as illustrated in Fig. \ref{fig: O-RAN}, in an O-RAN architecture, software modules known as xApps are deployed at near-real time (RT) radio intelligent controllers to run functions on behalf of disaggregated base stations. The AI-based ``apps'', such as the  xApps in O-RAN, should be tailored to the current network operating conditions, such as topology, traffic load, and channel quality, so as to meet given target performance levels\cite{O-RAN1,O-RAN2}.

The mapping between context and AI model is ideally done in a \emph{zero-shot} fashion by leveraging only contextual information, without requiring current data. As illustrated in Fig. \ref{fig: runtime}, this can be realized by means of an \emph{automatic model selection} (AMS) function that takes as input the context variables $c$, producing as its output the parameters $\phi=g(c)$ of an AI model\cite{AMS definition,AMS study1,AMS study2}. The AMS function $g(\cdot)$ may be a neural model, such as a transformer\cite{in-context}, while the model parameters $\phi$ may, for instance, be the weights of a graph neural network (GNN) used for power control (see Fig. \ref{fig:example})\cite{GNN1,GNN2,GNN3,GNN4}. 


However, before it can be deployed, the AMS mapping $g(\cdot)$ must be optimized during a preliminary phase, which is referred to as \emph{calibration} in Fig. \ref{fig: O-RAN}. This optimization phase generally requires collecting data from multiple contexts $c$ from the physical system, so as to be able to extrapolate the mapping $\phi=g(c)$ for new contexts $c$. Indeed, the physical system would have to be exposed to  many different network conditions, traffic statistics, and users' requests before the network operator can trust the AMS mapping to offer a reliable zero-shot recommendation of model parameters $\phi$ for a new context $c$.  Therefore, if carried out online, the initial AMS-mapping calibration phase  would be extremely time consuming. 

A possible solution to speed up calibration of the AMS mapping is to leverage a \emph{digital twin} (DT) of the physical system  to simulate data from multiple contexts. This way, calibration can be carried out, in principle, without requiring extensive real-world data. Digital twinning has been recently proposed as a paradigm for the design and monitoring of wireless systems\cite{DT advance1,DT advance2,DT advance3,DT advance4,DT advance5,DT advance6}. A key challenge in using DT-generated data is the inevitable real-to-sim gap caused by the discrepancy between DT simulation and the ground-truth data-generating mechanism. As a result, a direct use of simulated data for the optimization of the AMS mapping may yield poor performance when tested in the real system\cite{meta-learning study1,GNN1,DT advance5}. 

This paper proposes a novel method that corrects for the bias of the digital-twin simulator by means of limited real data collected from the physical system over a sequence of physical contexts. As shown in Fig. \ref{fig: calibration}, in the proposed calibration procedure, for each real-world context $c$, the digital twin produces synthetic data from multiple simulated contexts, and the real-world data is leveraged to ``rectify'' the errors made by the simulator\cite{PPI,PPI++}. The goal of the proposed online optimization of the AMS mapping is to reduce the time required by the calibration phase to produce an effective AMS mapping.

\subsection{Main  Contributions}
\begin{figure*}[htbp]
    \centering
    \includegraphics[width=11cm]{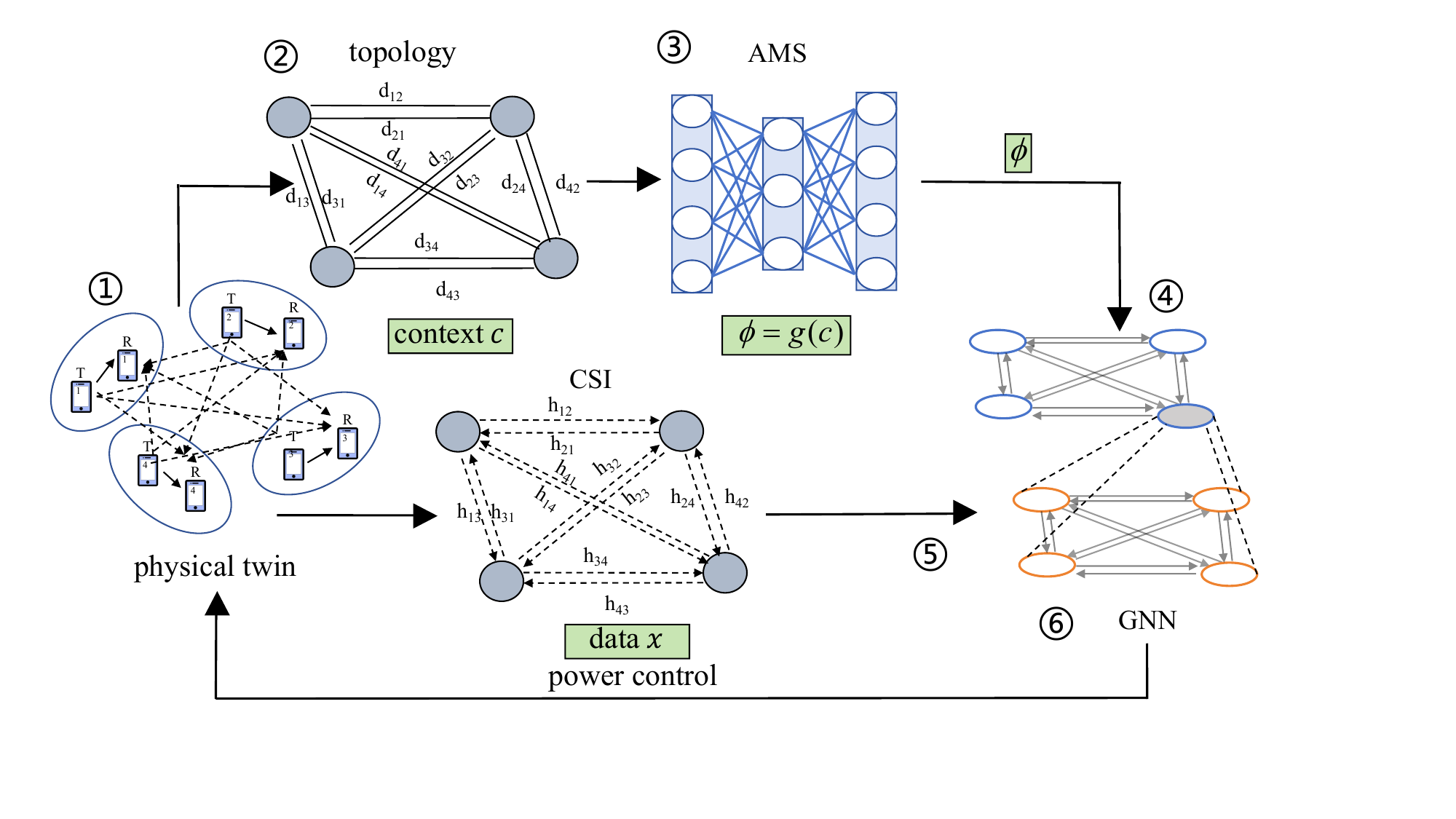}
    \vspace{-0.2cm}
    \caption{Wireless resource allocation using AMS: Given the context $c$ describing the matrix of geographical distances among the network nodes, i.e., the network topology, the AMS mapping $\phi=g(c)$ returns a GNN model $\phi$. The GNN model takes the current network CSI as input, and it outputs the transmission powers for all the nodes.}
    \label{fig:example}
    \vspace{-0.2cm}
\end{figure*}

This paper proposes a novel DT-powered online calibration scheme for the optimization of an AMS mapping.  As illustrated in Fig. \ref{fig:example}, the AMS mapping implements a zero-shot adaptation of the parameters $\phi$ of an AI model on the basis of a context variable $c$ that describes the task of interest. For example, the context information $c$ may encode the matrix of geographic distances obtained from the locations of the devices, and the AI model may be a GNN that determines a power allocation policy based on channel state information (CSI). 

A conventional online optimization of the AMS mapping, illustrated in Fig. \ref{fig:onlyPT}, would require the collection of a large amount of data from the physical system in order to obtain information for a sufficiently large number of contexts. The proposed approach, depicted in Fig. \ref{fig:PPI}, makes use of a simulator a DT in order to speed up the calibration process. The main focus of this work is addressing the inevitable mismatch between the synthetic data generated by the DT and the data that would have been generated in the same context by the real system. In order to overcome this challenge, we propose a novel online calibration strategy that leverages real data for the purpose of rectifying the  bias caused by the real-to-sim discrepancy. 

Our main contributions are as follows. 
\begin{itemize}
    \item We propose a novel \emph{DT-powered online AMS calibration} strategy that leverages the presence of a DT in the cloud. In order to rectify the real-to-sim discrepancy  between the DT simulator and the real-world data-generating distribution, we propose an optimization criterion that combines data from the current physical context and from multiple synthetic contexts. Specifically, inspired by the recently introduced statistical tool known as prediction-powered inference (PPI) \cite{PPI,PPI++}, we use synthetic data from the current physical context to estimate the DT simulation error. The estimated error is used to correct the optimization criterion evaluated by the DT based on synthetic data.
    \item In order to tailor PPI to the problem of calibrating AMS mappings, we introduce a novel \emph{adaptive} DT-powered online AMS calibration scheme that control the trade-off between bias and variance of the error-corrected optimization criterion. 
    \item Extensive experimental results for the problem of GNN-based power control\cite{GNN2} demonstrate the advantage of the proposed methodology in terms of convergence time for the online calibration phase. 
    
\end{itemize}

\subsection{Related Work}

\emph{Few-shot learning}: Meta-learning\cite{meta-learning} has been widely explored to adapt to dynamic wireless condition\cite{meta-learning study1, meta-learning study2, GNN1, GNN4}, assuming the accessibility of current data. In \cite{GNN4}, a modular meta-learning technique is proposed to accelerate the adaption by optimizing reusable modules. More recently, in-context learning via transformers was shown to provide an effective implementation for meta-learning in wireless systems\cite{in-context}.

\emph{Automatic AI model selection}: Automatic AI model selection refers to methods that automatically select an AI model for a given dataset\cite{AMS definition}. In \cite{AMS study1,AMS study2}, AI models are automatically selected for a wireless sensor network.

\emph{Digital twinning}: Digital twinning amounts to a virtual representation of a physical system that mirrors its physical attributes, processes, and dynamics. Digital twins support in what-if analysis\cite{what-if}, predictive simulations\cite{prediction}, and system optimization\cite{optimization1, optimization2}. Existing works in the communication area primarily focus on leveraging DT to optimize the network operation\cite{optimization1, optimization2, optimization3, DT advance4, optimization4, optimization5, optimization6}. For example, the authors of \cite{optimization1} have applied deep reinforcement learning (DRL) and transfer learning to propose a DT placement strategy; reference\cite{optimization2} introduced a sample-efficient beam pattern optimization based on DRL in a DT-based wireless system; and the work\cite{optimization3} proposed a resource scheduling method based on DRL and DT for unmanned aerial vehicle-assisted wireless systems. These works assume a perfect mirroring between physical system and the virtual representation. 

In contrast, reference\cite{DT advance7} formulated a joint optimization problem for the synchronization between the real and digital worlds in a mobile edge computing system, with the goal of minimizing the average synchronization time; and the authors of \cite{DT advance8} aimed at minimizing the overall transmission delay to ensure the synchronization of a digital twin-based wireless system.

\subsection{Organization}
The remainder of the paper is organized as follows. Section \ref{PT AMS}
presents a conventional AMS calibration strategy based only on real-world data. The proposed DT-powered AMS scheme and the improved adaptive version are respectively presented in Section \ref{DT AMS} and Section \ref{A DT AMS}. Section \ref{setup} describes the simulation setup, and Section \ref{simulation} illustrates the simulation results. Finally, Section \ref{conclusion} concludes the paper.

\section{Automatic AI Model Selection}\label{PT AMS}

In this section, we first define the objective of AMS, and then we review a baseline algorithm for the optimization of AMS that uses only data collected at the physical system. We summarize the main notations used throughout this paper in Table \ref{symbol_table}.

\begin{table*}[t]
    \renewcommand\arraystretch{1.2}
	\small
	\caption{Notation Used in This Paper}
	\label{symbol_table}
	\centering
	\scalebox{1}{
	\begin{tabular}{c|c}
		\toprule
		\textbf{Symbol} & \textbf{Definition}\\
		\hline\hline
		\makecell{$c$} & context variable\\
		\hline
        \makecell{$c_t$} & context variable collected from PT at time step $t$\\
        \hline
		\makecell{$c_{m,t}$} & $m$-th context variable generated by DT at time step $t$\\
		\hline
        \makecell{$\phi$} & parameter vector of the AI model \\
        \hline
        \makecell{$\theta$} & parameter vector of the AMS mapping \\
        \hline
        \makecell{$x$} & data\\
        \hline
        \makecell{$\ell(x|\phi)$} & loss accrued by model $\phi$ for data $x$\\
        \hline
        \makecell{$g(\cdot|\theta)$} & AMS mapping parameterized
        by a vector $\theta$\\
        \hline
        \makecell{$x_{n,t}$} & $n$-th data collected from PT at time step $t$\\
        \hline
        \makecell{$\tilde{x}_{n,t}$} & $n$-th data generated by DT at time step $t$\\
        \hline
        \makecell{$\tilde{x}_{m,n,t}$} & $n$-th data generated by DT based on context variable $c_{m,t}$\\
        \hline
        \makecell{$N^{\text{PT}}$} & number of data points collected from PT at each time step\\
        \hline
        \makecell{$N^{\text{DT}}$} & number of data points generated by DT at each time step\\
        \hline
        \makecell{$M^{\text{DT}}$} & number of context variables generated by DT at each time step\\
        \hline
        \makecell{$f$} & fidelity level of DT\\
        \hline
        \makecell{$S_f$} & simulation cost of DT at fidelity level $f$\\
        \hline
        \makecell{$S^{\text{DT}}$} & total simulation budget at DT\\
        \hline
        \makecell{$p(c)$} & distribution of context variable $c$\\
        \hline
        \makecell{$p(\cdot|c)$} & data generating distribution at PT given context $c$\\
        \hline
        \makecell{$p(\cdot|c,f)$} & simulated data-generating distribution $p(\cdot|c)$ at DT under fidelity given context $c$\\
        \hline
        
		\bottomrule
	\end{tabular}}
	\vspace{-0.1cm}
\end{table*}

\subsection{Automatic Model Selection}\label{AMS}
We study the problem of automatically selecting an AI model as a function of a context variable describing the task of interest. Formally, given a context variable $c$, the aim of AMS is to select an AI model described by a parameter vector $\phi$. To this end, AMS implements a mapping
\begin{equation}
    \phi=g(c)\label{mapping}
\end{equation}
between context $c$ and model $\phi$.
The focus of this paper is the design of the AMS mapping in (\ref{mapping}) using a data-driven methodology.

While our approach can be applied more broadly, we will use throughout the example illustrated in Fig. \ref{fig:example} for concreteness. In it, the context $c$ represents the topology of a wireless network, and the AI model to be selected is a graph neural network (GNN). The GNN takes CSI of the wireless network as an input, and it outputs the transmission powers for the nodes in the network. The works \cite{GNN2,GNN3} assumed a GNN model that is designed to perform well across a range of typologies, while the papers\cite{GNN1,GNN4} demonstrated the performance advantage accrued by adapting the model using data from the current topology via meta-learning. In this work, we aim at automatically mapping context $c$, i.e., topology, to GNN model $\phi$ using the AMS function (\ref{mapping}) without requiring data from the current topology. In this sense, AMS implements a form of zero-shot meta-learning\cite{zero-shot}.

Given a context $c$ at each time step, the performance of an AI model $\phi$ depends on a random vector $x\sim p(x|c)$, whose unknown distribution $p(x|c)$ is context-dependent. The vector $x$ includes the inputs to the AI model, as well as possibly other variables such as feedback signals. For example, for the setting in Fig. \ref{fig:example}, the vector $x$ is a realization of the CSI across all the links in the network. The distribution $p(x|c)$ is practically unknown, and it can be only approximated via modeling strategies, including ray tracing in the case of CSI\cite{ray-tracing}. We write as $\ell(x|\phi)$ the loss accrued by model $\phi$ for instance $x$, so that the average loss is given by
\begin{equation}
    \mathbb{E}_{x \sim p(x|c)}[\ell(x|\phi)].\label{loss definition}
\end{equation}

Given that the distribution $p(x|c)$ is unknown, the average loss in (\ref{loss definition}) can be only estimated using realizations of the random vector $x \sim p(x|c)$. However, in this paper, we assume that, at deployment time, no data $x \sim p(x|c)$ is available for the current context $c$. Rather, a model $\phi$ is selected via the AMS mapping (\ref{mapping}) based solely on the context variable $c$. Our goal is optimizing the AMS mapping (\ref{mapping}) during an initial calibration phase, so as to ensure a small value for the per-context loss (\ref{loss definition}).

\begin{figure*}[htbp]
    \centering
    \includegraphics[width=12cm]{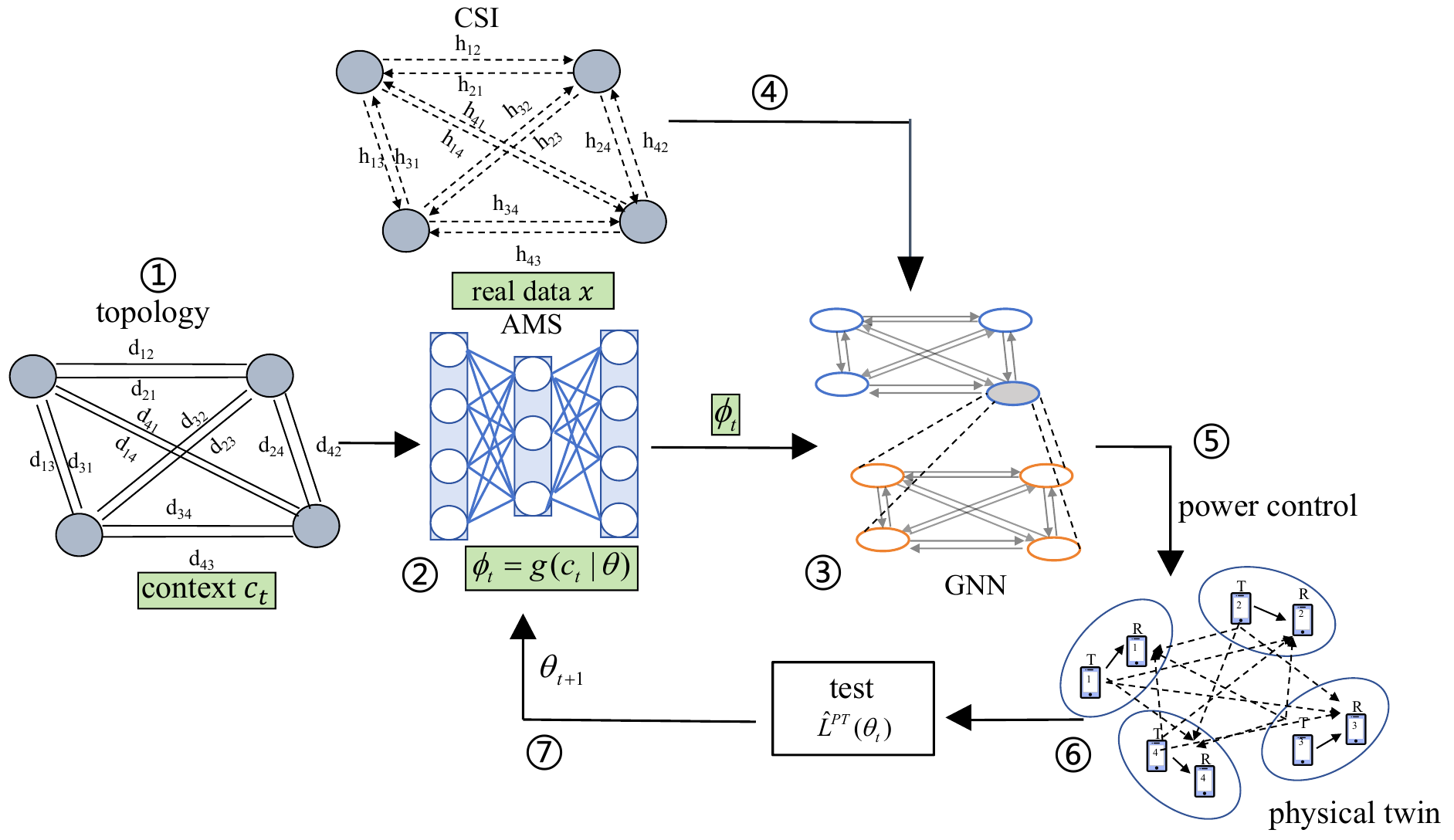}
    \vspace{-0.2cm}
    \caption{Conventional AMS calibration based on online learning: At each time step $t$, the physical twin (PT) is faced with a context $c_t\sim p(c)$, and it tests the current AMS output model $\phi_t=g(c_t|\theta_t)$ over $N^{\text{PT}}$ channel realizations, producing the estimate (\ref{PT loss}). Using online gradient descent, the AMS model parameter vector is updated using (\ref{onlyPT}).}
    \label{fig:onlyPT}
\end{figure*}

\subsection{Calibrating the AMS Mapping}
To enable the optimization of the AMS mapping, we adopt a function that is parameterized by a vector $\theta$, i.e.,
\begin{equation}
    \phi = g(c|\theta).
\end{equation}
Calibration of the AMS mapping $g(\cdot|\theta)$ aims at producing a parameter vector $\theta$ such that the model $\phi=g(c|\theta)$ yields a small value for the per-context loss in (\ref{loss definition}). More precisely, writing as $p(c)$ the marginal distribution of the context variable $c$, we are interested in minimizing the expected value of the per-context loss $\mathcal{L}(\phi|c)$ over the random context $c \sim p(c)$ as
\begin{equation}
\mathop{\rm{min}}\limits_{\theta} \left\{L^{{\text{PT}}}(\theta)=\mathbb{E}_{c\sim p(c)} \mathbb{E}_{x \sim p(x|c)}[\ell(x|g(c|\theta))]\right\}.\label{PT loss expectation}
\end{equation}
The superscript “PT” in the expected loss (\ref{PT loss expectation}) is used to emphasize that the loss $L^{{\text{PT}}}(\bm{\theta})$ is evaluated based on the real-world data-generating distribution $p(x|c)$, which pertains to physical twin (PT), i.e., to the real system.

During the calibration phase, at each discrete time step $t=1,2,...$, the system is faced with a context $c_t\sim p(c)$. Note that the context variables $c_t$ may be correlated across the time index $t = 1,2,\dots$, as long as they share the same marginal distribution $p(c)$. For example, the sequence of context variables $c_1,c_2,\dots$, may follow a Markov model with stationary distribution $p(c)$. By testing the current AMS model parameter $\theta_t$, the system estimates the PT loss $L^{{\text{PT}}}(\theta_t)$ via the empirical average
\begin{equation}
     \hat{L}^{\text{PT}}({\theta}_t) = \frac{1}{N^{{\text{PT}}}}\sum_{n=1}^{N^{{\text{PT}}}}\ell(x_{n,t}|g(c_{t}|\theta_t))\label{PT loss}
\end{equation}
over i.i.d. realizations $\{x_{n,t}\}_{n=1}^{N^{\text{PT}}}$ from the distribution $p(x|c)$. For example, in the setting of Fig. \ref{fig:onlyPT}, the estimate (\ref{PT loss}) is obtained by testing the current power-allocation GNN $\phi_t=g(c_t|\theta_t)$ at $N^{\text{PT}}$ different CSI realizations $\{x_{n,t}\}_{n=1}^{N^{\text{PT}}}$. Note that, while the empirical loss function $\hat{L}^{\text{PT}}(\cdot)$ itself depends on time index $t$, we omit the subscript $t$ to avoid clutter. Furthermore, the estimate (\ref{PT loss}) is unbiased, i.e., 
\begin{equation}
    \mathbb{E}[\hat{L}^{\text{PT}}(\theta_t)]=L^{\text{PT}}(\theta_t),
\end{equation}
where the expectation is taken over the data $\{x_{n,t}\}_{n=1}^{N^{\text{PT}}}\mathop{\sim}\limits^{\text{i.i.d.}}p(x|c_t)$ as well as the context $c_t \sim p(c)$.

Apart from the empirical loss, we assume that the physical system can also obtain the gradient $\nabla\hat{L}^{\text{PT}}({\theta}_t)$. This can be estimated, for instance, using a perturbation-based approach\cite{gradient-estimation}. With this information, the AMS parameter vector $\theta_t$ can be updated in an online fashion as
\begin{equation}
    \theta_{t+1} = \theta_{t}-\gamma_t\nabla\hat{L}^{\text{PT}}(\theta_t),\label{onlyPT}
\end{equation}
where $\gamma_t > 0 $ is a step size at time $t$. This approach is illustrated in Fig. \ref{fig:onlyPT}.

The conventional calibration scheme (\ref{onlyPT}) has the important drawback that testing the AMS mapping under different contexts $c_t$ is time consuming, limiting the number $N^{\text{PT}}$ of data points used in (\ref{PT loss}) and the overall number of steps $t$ that can be allocated to AMS calibration. The proposed approach, to be introduced in the next section, tackles this problem by resorting synthetic data produced by a simulator in a DT.

\section{DT-powered Automatic AI Model Selection}\label{DT AMS}
In this section, we introduce the proposed AMS calibration scheme that leverages a DT to accelerate the convergence to a good approximation of the optimization problem (\ref{PT loss expectation}). To this end, we first describe \emph{na\"ive DT-powered AMS} (N-DT-AMS), a baseline approach that treats synthetic data on par with the real-world data produced by the PT. Then, we introduce the proposed \emph{DT-powered AMS} (DT-AMS), which uses real-world data to correct the bias caused by the discrepancy between the DT simulator and the real-world data generating distribution.

\subsection{Digital Twin}
Following the conventional definition of digital twining\cite{DT_definition}, in this paper, the DT is a simulator of the PT (see also \cite{optimization2, optimization4, DT advance1, DT advance3}). Specifically, given a context variable $c$, the DT can generate synthetic data at some fidelity level $f \in \{1,\dots,F_{\rm{max}}\}$. At fidelity level $f$, the synthetic data $\tilde{x}$ generated by the DT follows the distribution $p(\cdot|c,f)$, which is an approximation of the true context-dependent distribution $p(\cdot|c)$. The quality of the approximation $p(\tilde{x}|c,f)$ improves as the fidelity $f$ increases, but a larger fidelity $f$ entails a larger simulation cost $S_f$\cite{fidelity}.  Accordingly, the simulation costs are ordered as
\begin{equation}
    S_1 \leq S_2 \leq \dots \leq S_{F^{\rm{max}}}.
\end{equation}

At each time step $t=1,2,\dots$ of the calibration phase, the DT simulates the system for a number $M^{\text{DT}}$ context variables $\{c_{m,t}\}_{m=1}^{M^{\text{DT}}}$ following the marginal distribution $p(c)$. Specifically, given a fidelity level $f$, the $N^{\text{DT}}$ data points generated at each context $c_{m,t}$ are distributed as
\begin{equation}
    \tilde{x}_{n,m,t}\mathop{\sim}\limits^{\text{i.i.d.}}p(\tilde{x}|c_{m,t},f)
\end{equation}
for $n=1,\dots,N^{\text{DT}}$. Assuming a total simulation budget given by $S^\text{DT}$, the number of context variables $M^{\text{DT}}$, the number of data points per context $N^{\text{DT}}$, and the fidelity level $f$ must satisfy the inequality
\begin{equation}
    S_fM^{\text{DT}}N^{\text{DT}}\leq S^\text{DT}.\label{budget constraint}
\end{equation}
In this regard, increasing the fidelity $f$ and thus the cost $S_f$, reduces the discrepancy between DT model $p(\cdot|c,f)$ and true distribution $p(\cdot|c)$, but at the expense of reducing the overall number $M^{\text{DT}}N^{\text{DT}}$ of synthetic data points that can be generated by the DT. We assume that the fidelity level $f$ is kept constant for the entire calibration process.

Using synthetic data, the DT can estimate the DT loss $L^{\text{DT}}(\theta_t)$ at the current AMS parameter vector $\theta_t$ as
\begin{equation}
    \hat{L}^{\text{DT}}(\theta_t) = \frac{1}{M^{\text{DT}}}\sum_{m=1}^{M^{\text{DT}}}\frac{1}{N^{\text{DT}}}\sum_{n=1}^{N^{\text{DT}}}\ell(\tilde{x}_{m,n,t}|g(c_{m,t}|\theta_t)).\label{DT loss tilde} 
\end{equation}
As long as the context variables follow the marginal, stationary, distribution $p(c)$, the average over the DT estimate (\ref{DT loss tilde}) is
\begin{equation}
    \mathbb{E}[\hat{L}^{\text{DT}}(\theta_t)] = L^{\text{DT}}(\theta_t),
\end{equation}
where we have defined the expected DT loss as
\begin{equation}
    L^{{\text{DT}}}(\theta) =\mathbb{E}_{c\sim p(c), \tilde{x} \sim p(\tilde{x}|c,f)}[\ell(\tilde{x}|g(c|\theta))].\label{DT loss expectation}
\end{equation}

Note that the assumption that the DT is aware of the stationary context distribution $p(c)$ is well justified in settings, such as in Fig. 2, in which the distribution $p(c)$ reflects macroscopic long-term trends, including the spatial density of devices \cite{GNN2,GNN3,GNN5}. That said, as further discussed in the next section, it is also possible to account for differences between the true distribution $p(c)$ and the sampling distribution assumed by the DT. Furthermore, the contexts $\{c_{m,t}\}_{m=1}^{M^{\text{DT}}}$ generated by the DT may potentially be correlated in order to capture the memory of the real context process. This is elaborated on in Section \ref{markov_simulation}. 

Importantly, due to the discrepancy between DT distribution $p(\cdot|c,f)$ and the real-world distribution $p(\cdot|c)$, the expected DT loss (\ref{DT loss expectation}) is generally different from the PT loss $L^{{\text{PT}}}(\theta)$ in (\ref{PT loss expectation}), i.e., 
\begin{equation}
   L^{{\text{DT}}}(\theta) \neq L^{{\text{PT}}}(\theta),\label{neq}
\end{equation}
and thus the DT loss in (\ref{DT loss tilde}) is a biased estimate of the expected PT loss (\ref{PT loss expectation}).

\subsection{Na\"ive DT-Powered AMS}
The simplest way for the DT to use both synthetic and real-world data is to weigh both sources of data equally, that is, by treating the synthetic data as if it had been genuine real-world data. Despite its simplicity, this approach plays an important role in semi-supervised learning literature as a strong baseline (see, e.g., \cite{initial choice}).

Using the empirical losses (\ref{PT loss}) and (\ref{DT loss tilde}), this yields to the estimate
\begin{small}
\begin{equation}
L^{{\text{N-DT-AMS}}}(\theta_t)=\frac{N^{\text{PT}}}{N^{\text{PT}}+M^{\text{DT}}N^{\text{DT}}}\hat{L}^{\text{PT}}(\theta_t)+\frac{M^{\text{DT}}N^{\text{DT}}}{N^{\text{PT}}+M^{\text{DT}}N^{\text{DT}}}\hat{L}^{\text{DT}}(\theta_t)\label{naive loss}
\end{equation}
\end{small}
at time step $t$. The estimate in (\ref{naive loss}) is biased with respect to the PT loss in (\ref{PT loss expectation}), although it may have a smaller variance as compared to the estimate $\hat{L}^{\text{PT}}(\theta_t)$ in (\ref{PT loss}). Based on (\ref{naive loss}), N-DT-AMS updates the AMS model parameter vector as
\begin{equation}
    \theta_{t+1} = \theta_{t}-\gamma_t\nabla L^{{\text{N-DT-AMS}}}(\theta_t),\label{naive}
\end{equation}
where $\gamma_t>0$ is the step size at time $t$, and $\nabla L^{{\text{N-DT-AMS}}}(\theta_t)$ is the gradient of loss $L^{{\text{N-DT-AMS}}}(\theta_t)$.

\subsection{DT-Powered AMS}
The N-DT-AMS loss estimate (\ref{naive loss}) can be significantly biased due to the discrepancy between the DT distribution $p(\cdot|c,f)$ and the real-world distribution $p(\cdot|c)$. Based on this observation, the proposed DT-AMS scheme uses synthetic data not only to augment the available data as in N-DT-AMS, but also to quantify, and partially correct, the mismatch between distributions $p(\cdot|c,f)$ and $p(\cdot|c)$. 

The key idea is that the DT uses data simulated for the current context $c_t$ in order to evaluate the bias caused by simulation errors. To this end, at each time step $t$ of the calibration phase, the PT informs the DT about the current context variable $c_t$. With this information, the DT simulates data $\{\tilde{x}_{n,t}\}_{n=1}^{N^{\text{DT}}}$ with $\tilde{x}_{n,t}\mathop{\sim}\limits^{\text{i.i.d.}}p(\tilde{x}|c_t,f)$ under the current context $c_t$ for the selected fidelity level $f$. If the simulation budget in (\ref{budget constraint}) allows, the $M^{\text{DT}}-1$ remaining contexts $\{c_{m,t}\}_{m=1}^{M^{\text{DT}}-1}$ are chosen from the marginal distribution $p(c)$ as in N-DT-AMS.
Using the synthetic data $\{\tilde{x}_{n,t}\}_{n=1}^{N^{\text{DT}}}$ generated for the current context $c_t$, the DT estimates the average loss as
\begin{equation}\label{DT loss}
   \hat{L}^{\text{DT}\rightarrow\text{PT}}(\theta_t) = \frac{1}{N^{\text{DT}}}\sum_{n=1}^{N^{\text{DT}}}\ell(\tilde{x}_{n,t}|g(c_t|\theta_t)).
\end{equation}
This estimate can be directly compared with the estimate $\hat{L}^{\text{PT}}(\theta_t)$ in (\ref{PT loss}), which is obtained from real-world data under the same context $c_t$. Accordingly, the DT estimates the bias caused by the simulator as the difference
\begin{equation}\label{correction term}
    \Delta\hat{L}^{\text{DT}\rightarrow\text{PT}}(\theta_t) = \hat{L}^{\text{PT}}(\theta_t) - \hat{L}^{\text{DT}\rightarrow\text{PT}}(\theta_t).
\end{equation}

Having estimated the DT-to-PT bias in (\ref{correction term}), the DT adds this correction to the loss $\hat{L}^{\text{DT}}(\theta_t)$ estimated using the remaining simulation budget as 
\begin{equation}
L^{{\text{DT-AMS}}}(\theta_t)= \hat{L}^{\text{DT}}(\theta_t)+\Delta\hat{L}^{\text{DT}\rightarrow\text{PT}}(\theta_t),\label{ppi loss}
\end{equation}
where $\hat{L}^{\text{DT}}(\theta_t)$ is the estimate obtained using the $M^{\text{DT}}-1$ contexts $\{c_{m,t}\}_{m=1}^{M^{\text{DT}}-1}$ in a manner similar to (\ref{DT loss tilde}), i.e.,
\begin{equation}
    \hat{L}^{\text{DT}}(\theta_t):= \frac{1}{M^{\text{DT}}-1}\sum_{m=1}^{M^{\text{DT}}-1}\frac{1}{N^{\text{DT}}}\sum_{n=1}^{N^{\text{DT}}}\ell(\tilde{x}_{m,n,t}|g(c_{m,t}|\theta_t)). \label{DT loss tilde ppi} 
\end{equation}

The DT-AMS empirical loss $L^{{\text{DT-AMS}}}(\theta_t)$ in (\ref{ppi loss}) is an unbiased estimate of the PT loss $L^{{\text{PT}}}(\theta_t)$, since the expectation over the context $c\sim p(c)$ and over real and synthetic data yields the equality
\begin{equation}\label{ppi expectation}
   \mathbb{E}[L^{{\text{DT-AMS}}}(\theta_t)]=L^{{\text{DT}}}(\theta_t)+L^{{\text{PT}}}(\theta_t)-L^{{\text{DT}}}(\theta_t)=L^{{\text{PT}}}(\theta_t).
\end{equation}
In fact, by construction, the expected value of the estimated loss (\ref{DT loss}) equals $\mathbb{E}[\hat{L}^{\text{DT}\rightarrow\text{PT}}(\theta_t)] = L^{\text{DT}}(\theta_t)$. 

\begin{figure*}[htbp]
    \centering
    \includegraphics[width=14cm]{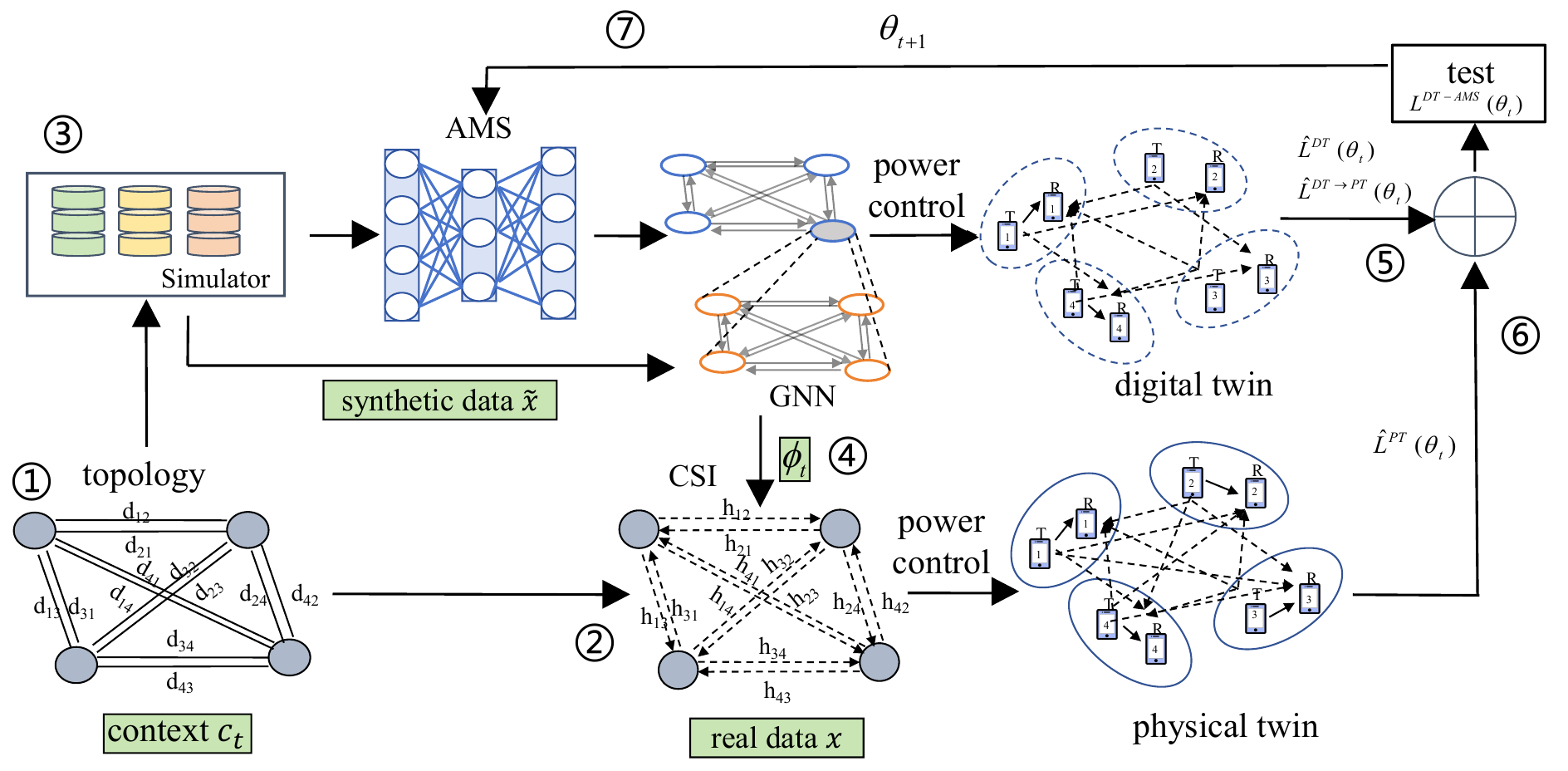}
    \vspace{-0.2cm}
    \caption{DT-powered AMS calibration based on online learning: At each time step $t$, the PT is faced with a context $c_t\sim p(c)$, and it collects corresponding real-world data $x_{n,t}\sim p(x|c_t)$ for $n=1,\dots,N^{\text{PT}}$. The PT sends the current context $c_t$ to the DT, which generates corresponding simulated data $\tilde{x}_{n,t}\sim p(\tilde{x}|c_t,f)$ for $n=1,\dots,N^{\text{DT}}$, as well as data $\tilde{x}_{m,n,t}\sim p(\tilde{x}|c,f)$ for $n=1,\dots,N^{\text{DT}}$ under $M^{\text{DT}}-1$ independent context variables $c_{m,t}\sim p(c)$ for $m=1,\dots,M^{\text{DT}}-1$. The DT uses the synthetic data points to evaluate the empirical losses (\ref{DT loss}) and (\ref{DT loss tilde ppi}). Upon reception of the model $\phi_t$ from the DT, the PT runs it over $N^{\text{PT}}$ channel realizations, producing the  loss estimate (\ref{PT loss}). Using online gradient descent, the AMS model parameter vector is updated using (\ref{ppi}).}
    \label{fig:PPI}
\end{figure*}

DT-AMS applies online gradient descent on the loss (\ref{ppi loss}) to update the AMS model parameter vector as 
\begin{equation}
    \theta_{t+1} = \theta_{t}-\gamma_t\nabla L^{{\text{DT-AMS}}}(\theta_t),\label{ppi}
\end{equation}
where $\gamma_t >0$ is the step size at time $t$. The overall approach is illustrated in Fig. \ref{fig:PPI}.

We note that the DT-AMS loss (\ref{ppi loss}) is inspired by prediction-powered inference (PPI)\cite{PPI,PPI++}. PPI addresses an offline estimation problem under a convex loss, for which (asymptotic) theoretical properties are established in\cite{PPI,PPI++}. In contrast, DT-AMS operates online, and it incorporates aspects specific to the considered DT-based scenarios such as the fidelity of the simulation. Unlike \cite{PPI,PPI++}, we allow for arbitrary losses, hence supporting any AMS function in (\ref{mapping}), and we do not address asymptotic theoretical properties.

\section{Adaptive DT-powered Automatic AI Model Selection}\label{A DT AMS}
In this section, we introduce an improved version of DT-AMS, which judiciously adapts the use of the synthetic data generated by the DT to the quality of the simulation. 

\subsection{Adaptive DT-Powered AMS}
As discussed in the previous section, the empirical loss (\ref{ppi loss}) adopted by DT-AMS is an unbiased estimate of the PT loss (\ref{PT loss expectation}). However, enforcing unbiasedness may be counterproductive, particularly in the early stages of calibration. In fact, it was observed that, at the first iterations of stochastic gradient-based optimization, it can be advantageous to trade some bias for a lower variance of the gradients \cite{var-bias,variance important}. The proposed approach introduces new  hyperparameters in the loss (\ref{ppi loss}) with the goal of optimally controlling the trade-off between bias and variance.

Specifically, at each time step $t$, adaptive DT-AMS (A-DT-AMS) adopts the loss
\begin{equation}
   L^{\text{A-DT-AMS}}(\theta_t) = \lambda_t \hat{L}^{\text{DT}}(\theta_t)+ \left(\hat{L}^{\text{PT}}(\theta_t)-\mu_t \hat{L}^{\text{DT}\rightarrow \text{PT}}(\theta_t)\right), \label{proposed loss}
\end{equation}
with hyperparameters $\lambda_t \geq 0$ and $\mu_t \geq 0$. The DT-AMS loss (\ref{ppi loss}) is recovered with $\lambda_t=\mu_t=1$, while the baseline PT-only scheme (\ref{PT loss}) is obtained with $\lambda_t = \mu_t = 0$. 

A-DT-AMS applies online gradient descent using the loss (\ref{proposed loss}) to update the AMS model parameter vector as 
\begin{equation}
    \theta_{t+1} = \theta_{t}-\gamma_t\nabla L^{{\text{A-DT-AMS}}}(\theta_t),\label{proposed}
\end{equation}
where $\gamma_t >0$ is the step size at time $t$. As discussed in the next subsection, the hyperparameters $\lambda_t$ and $\mu_t$ are automatically tuned as a function of time step $t$ with the aim of 
minimizing the mean squared error (MSE) in the estimate of the loss $L^{\text{PT}}(\theta)$.

\subsection{Adapting the Hyperparameters}
The MSE of the loss (\ref{proposed loss}) with respect to the PT loss (\ref{PT loss expectation}) can be evaluated as 
\begin{equation}\label{MSE}
\begin{aligned}
    \mathrm{MSE}(\theta_t) &= \mathbb{E}\left[\left(L^{\text{A-DT-AMS}}(\theta_t) - L^\text{PT}(\theta_t)\right)^2\right].
    \end{aligned}
\end{equation}
We propose to tune the hyperparameters $\lambda_t$ and $\mu_t$ so as to minimize the MSE in (\ref{MSE}), i.e., 
\begin{equation}\label{lambda mu problem}
   \{\lambda_t^*,\mu_t^*\} =  \mathop{\rm{argmin}}\limits_{\{\lambda_t,\mu_t\}}\ \mathrm{MSE}(\theta_t). 
\end{equation}
As shown in Appendix \ref{lambda mu solution}, for any AMS model parameters $\theta_t$, we have an explicit solution for the minimization (\ref{lambda mu problem}) as
\begin{equation}\label{lambda mu solution text}
        \begin{aligned}
            \lambda_t^* &= \frac{C^{\text{DT}\rightarrow\text{PT}}(\theta_t){L^{\text{DT}}(\theta_t)}^2}{{L^{\text{DT}}(\theta_t)}^2\left(V^{\text{DT}\rightarrow \text{PT}}(\theta_t)+V^{\text{DT}}(\theta_t)\right)+V^{\text{DT}\rightarrow \text{PT}}(\theta_t)V^{\text{DT}}(\theta_t)},\\
            \mu_t^* &= \frac{C^{\text{DT}\rightarrow\text{PT}}(\theta_t)+\lambda_t^*{L^{\text{DT}}(\theta_t)}^2}{V^{\text{DT}\rightarrow \text{PT}}(\theta_t)+{L^{\text{DT}}(\theta_t)}^2},
        \end{aligned}
    \end{equation}
where $V^{\text{DT}}(\theta_t)$, and $V^{\text{DT}\rightarrow \text{PT}}(\theta_t)$ are defined as the variance of the estimates  $\hat{L}^{\text{DT}}(\theta_t)$ in (\ref{DT loss tilde ppi}), and $\hat{L}^{\text{DT}\rightarrow\text{PT}}(\theta_t)$ in (\ref{DT loss}), respectively, i.e., 
\begin{align}
    V^{\text{DT}}(\theta_t) &= \mathbb{E}[(\hat{L}^{\text{DT}}(\theta_t) - L^{\text{DT}}(\theta_t))^2],\label{VDT tilde}\\
    V^{\text{DT}\rightarrow \text{PT}}(\theta_t) &= \mathbb{E}[(\hat{L}^{\text{DT}\rightarrow \text{PT}}(\theta_t) - L^{\text{DT}}(\theta_t))^2]; \label{VDT} 
\end{align}
while $C^{\text{DT}\rightarrow \text{PT}}(\theta_t)$ is defined as the covariance between the estimates $\hat{L}^{\text{PT}}(\theta_t)$ and $\hat{L}^{\text{DT}\rightarrow\text{PT}}(\theta_t)$, i.e.,
\begin{equation} \label{C_PT_DT}
    C^{\text{DT}\rightarrow \text{PT}}(\theta_t) = \mathbb{E}\left[\left(\hat{L}^{\text{PT}}(\theta_t) - L^{\text{PT}}(\theta_t)\right)\left(\hat{L}^{\text{DT}\rightarrow\text{PT}}(\theta_t) - L^{\text{DT}}(\theta_t)\right)\right].  
\end{equation} 
The optimal values (\ref{lambda mu solution text}) rely on knowledge at the DT about the marginal context distribution $p(c)$. As shown in Appendix \ref{lambda mu solution}, it is possible to generalize the optimal solution (\ref{lambda mu solution text}) to a setting in which the DT only has an available estimate $\hat{p}(c)$ of the true context distribution $p(c)$.

\subsection{Online Estimation of the Hyperparameters}
The variances and covariance in (\ref{VDT tilde})-(\ref{C_PT_DT}), which are required to evaluate the optimal hyperparameters in (\ref{lambda mu solution text}), cannot be computed, as one does not have access to the true underlying distribution. To obviate this problem, in this subsection, we propose a practical moving-average estimation scheme to obtain the values of $\hat{\lambda}_t^*$ and $\hat{\mu}_t^*$.

Specifically, we estimate $V^{\text{DT}}(\theta_t)$, $V^{\text{DT}\rightarrow \text{PT}}(\theta_t)$, and $C^{\text{DT}\rightarrow \text{PT}}(\theta_t)$ via the moving averages
\begin{subequations}\label{estimate v and c}
    \begin{align}
    \hat{V}^{\text{DT}}(\theta_t) &= \frac{1}{W-1}\sum_{t^{'}=t-W+1}^t\left(\hat{L}^{\text{DT}}(\theta_{t^{'}})- \hat{\hat{L}}^{\text{DT}}(\theta_{t})\right)^2,\\
    \hat{V}^{\text{DT}\rightarrow\text{PT}}(\theta_t) &= \frac{1}{W-1}\sum_{t^{'}=t-W+1}^t\left(\hat{L}^{\text{DT}\rightarrow\text{PT}}(\theta_{t^{'}})- \hat{\hat{L}}^{\text{DT}\rightarrow\text{PT}}(\theta_{t})\right)^2,\\
    \hat{C}^{\text{DT}\rightarrow \text{PT}}(\theta_{t}) &= \frac{1}{W-1}\sum_{t^{'}=t-W+1}^t\left(\hat{L}^{\text{PT}}(\theta_{t^{'}})-\hat{\hat{L}}^{\text{PT}}(\theta_{t})\right)\left(\hat{L}^{\text{DT}\rightarrow \text{PT}}(\theta_{t^{'}})-\hat{\hat{L}}^{\text{DT}\rightarrow \text{PT}}(\theta_{t})\right),\label{CPT-DT}
\end{align}
\end{subequations}
over the previous $W$ time steps. In (\ref{estimate v and c}), the quantities $\hat{\hat{L}}^{\text{DT}}(\theta_{t})$, $\hat{\hat{L}}^{\text{DT}\rightarrow\text{PT}}(\theta_{t})$, and $\hat{\hat{L}}^{\text{PT}}(\theta_{t})$ represent the average of the estimates $\hat{L}^{\text{DT}}(\theta_{t})$, $\hat{L}^{\text{DT}\rightarrow\text{PT}}(\theta_{t})$, and $\hat{L}^{\text{PT}}(\theta_{t})$, respectively, over the past $W$ time steps, i.e.,
\begin{subequations}\label{loss average}
    \begin{align}
    \hat{\hat{L}}^{\text{DT}}(\theta_{t}) &= \frac{1}{W}\sum_{t^{''}=t-W+1}^t\hat{L}^{\text{DT}}(\theta_{t^{''}}),\\
    \hat{\hat{L}}^{\text{DT}\rightarrow\text{PT}}(\theta_{t}) &= \frac{1}{W}\sum_{t^{''}=t-W+1}^t\hat{L}^{\text{DT}\rightarrow\text{PT}}(\theta_{t^{''}}),\\
    \hat{\hat{L}}^{\text{PT}}(\theta_{t}) &= \frac{1}{W}\sum_{t^{''}=t-W+1}^t\hat{L}^{\text{PT}}(\theta_{t^{''}}).
\end{align}
\end{subequations}
Using the estimates $\hat{V}^{\text{DT}}(\theta_t)$, $\hat{V}^{\text{DT}\rightarrow \text{PT}}(\theta_t)$, $\hat{C}^{\text{DT}\rightarrow \text{PT}}(\theta_{t})$, and $\hat{\hat{L}}^{\text{DT}}(\theta_{t})$ in lieu of $V^{\text{DT}}(\theta_t)$, $V^{\text{DT}\rightarrow \text{PT}}(\theta_t)$, $C^{\text{DT}\rightarrow \text{PT}}(\theta_t)$, and $L^{\text{DT}}(\theta_{t})$, respectively, we estimate the optimal parameters $\hat{\lambda}_t^*$ and $\hat{\mu}_t^*$ based on (\ref{lambda mu solution text}). In order to avoid transient issues, in practice, we propose to fix the hyperparameters $\lambda_t$ and $\mu_t$ to some set values $\lambda_0$ and $\mu_0$ for the first $W$ time instants. The proposed A-DT-AMS scheme is summarized in Algorithm \ref{A1}.

\begin{figure*}[bpth]
    \centering
    \includegraphics[width=13cm]{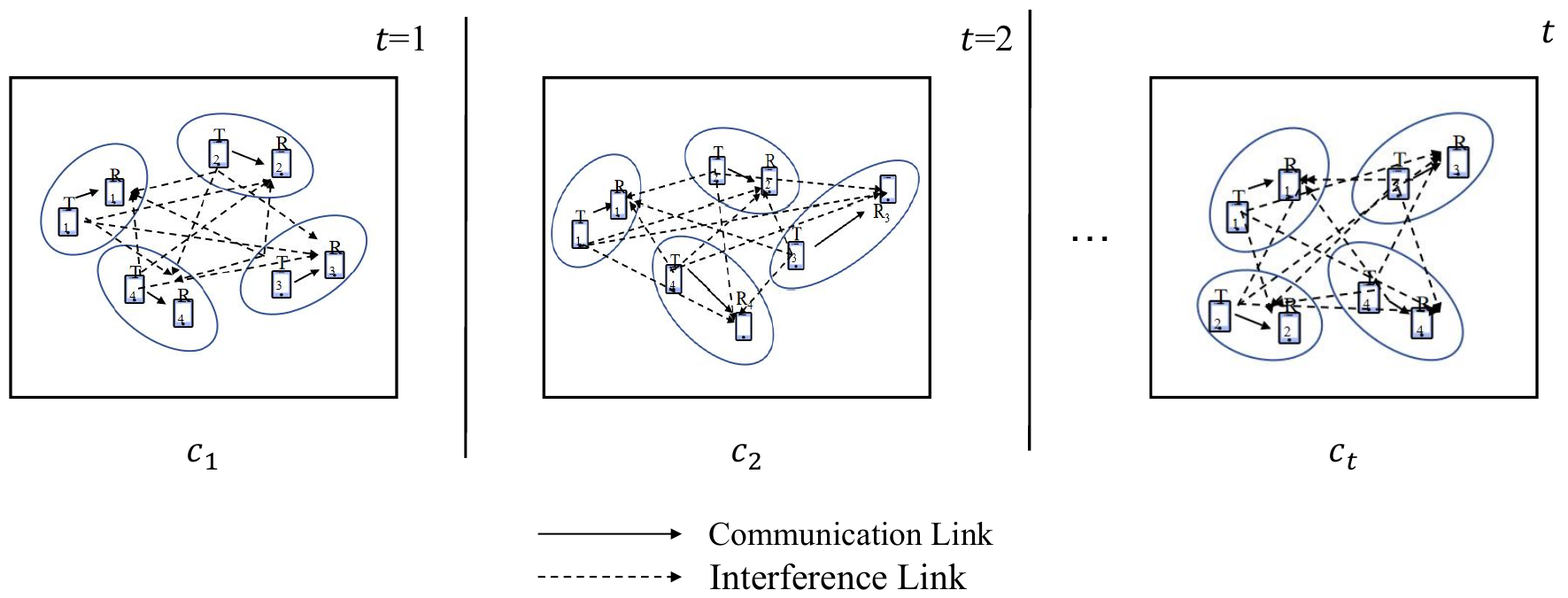}
    \caption{Time-varying $K$-user interference network assumed in the experimental results.}
    \label{fig:system model}
\end{figure*}

\begin{table}[htp]
	\setlength{\abovecaptionskip}{-2pt}
	\setlength{\belowcaptionskip}{-6pt}
    \setstretch{1}
	\begin{algorithm}[H]
		\caption{\textbf{Adaptive DT-Powered AMS Calibration}}
		\label{A1}
		{\normalsize
			\begin{algorithmic}[1]
               \State \textbf{Input:} Total simulation budget $S^{\text{DT}}$, fidelity level $f$, initial hyperparameters $\lambda_0$ and $\mu_0$, and initial learning rate in (\ref{proposed}), $\gamma_0$
               \State \textbf{Output:} AMS mapping $g(\cdot|\theta)$
               \For{$t=1,2,\dots,T$}
               \State PT collects real-world data $x_{n,t}\sim p(x|c_t)$ for $n=1,\dots,N^{\text{PT}}$ for current context $c_t \sim p(c)$ 
               \State PT sends $c_t$ to DT 
               \State DT generates simulated data $\tilde{x}_{n,t}\sim p(\tilde{x}|c_t,f)$ for $n=1,\dots,N^{\text{DT}}$, as well as data $\tilde{x}_{m,n,t}\sim p(\tilde{x}|c,f)$ for $n=1,\dots,N^{\text{DT}}$ for context variables $c_{m,t}\sim p(c)$, $m=1,\dots,M^{\text{DT}}-1$ 
               \State DT sends $\phi_t$ to PT
               \State PT calculates the empirical PT loss (\ref{PT loss})
               \State PT sends the empirical PT loss (\ref{PT loss}) and its gradient to DT
               \State DT calculates the empirical DT loss (\ref{DT loss}) and (\ref{DT loss tilde ppi}) 
               \If{$t \textless W$}
               \State DT sets ${\lambda}_t=\lambda_0$ and ${\mu}_t=\mu_0$
               \Else
               \State DT calculates $\lambda_t$ and $\mu_t$ based on (\ref{lambda mu solution text})-(\ref{loss average})
               \EndIf
               \State Using gradient update, the AMS model parameter vector $\theta_t$ is updated via (\ref{proposed})
               \EndFor
            \end{algorithmic}}
	\end{algorithm}
 \vspace{-0.5cm}
\end{table}

\section{Simulation Setup} \label{setup}
In this section, we describe the setting used in the next section to compare different AMS strategies. We focus on the setup illustrated in Fig. \ref{fig:system model}, in which an AI app based on a GNN is used to implement a power allocation strategy in a network consisting of multiple inferring links. We first describe the setup and then cover implementation aspects such as the neural model used for the AMS mapping.

\subsection{Setup}\label{small set-up}
\subsubsection{Scenario}

As depicted in Fig. \ref{fig:system model}, following a common formulation of the power control problem in the presence of multiple transceiver pairs\cite{GNN2,GNN3}, the PT consists of a network with $K$ single-antenna transceiver pairs $\{(T_t^k,R_t^k)\}^K_{k=1}$ located in a square deployment area with sides of length $A$. We denote as $T_t^k$ and $R_t^k$ the positions of the $k$-th transmitter and receiver, respectively, at each time step $t$ of the calibration phase. At time step $t$, the position $T_t^k$ of the $k$-th transmitter is randomly and uniformly selected in the deployment area, while the corresponding receiver's position $R_t^k$ is randomly and uniformly chosen within a two-dimensional torus centered at $T_t^k$ with radii $d_{\rm{min}}$ and $d_{\rm{max}}$.

Given the position $\{(T_t^k,R_t^k)\}^K_{k=1}$, the pair-wise distances between each transmitter and all the receivers at each time step $t$ are collected in the asymmetric distance matrix $\mathbf{D}_t\in \mathbb{R}^{K\times K}$, in which each $(j,k)$-th entry $d^{j,k}_t$ is the distance between the $j$-th transmitter and the $k$-th receiver at time step $t$. We denote as $p(\mathbf{D})$ the distribution over the distance matrix $\mathbf{D}$ produced by the described distribution of the transmitters and receivers. The distance matrix $\mathbf{D}_t$ describes the context $c_t$.

Given the distance matrix $\mathbf{D}_t$, the complex channel coefficient $h^{j,k}_t\in\mathbb{C}$ between the transmitter $j$ and the receiver $k$ at time step $t$ accounts for large-scale fading and small-scale fading as \cite{large-scale, small-scale} 

\begin{equation}\label{Rician fading}
    h^{j,k}_t = 10^{-L(d^{j,k}_t)/20}\sqrt{\eta\sigma_s^2}\cdot \left(\sqrt{\frac{\kappa}{\kappa+1}}e^{j\alpha_{j,k}} + \sqrt{\frac{1}{\kappa+1}}\hat{h}_{j,k}\right),
\end{equation}
where $L(d^{j,k}_t)$ is the path-loss at distance $d^{j,k}_t$ \cite{large-scale}; $\eta$ is the transmit antenna power gain; $\sigma_s^2$ is the shadowing power; $\alpha_{j,k} \sim \mathcal{U}[-\pi, \pi]$ is the random phase of the line-of-sight (LoS) component; $\hat{h}_{j,k}\sim \mathcal{CN}(0,1)$ models Rayleigh fading for the non line-of-sight (NLoS) component; and $\kappa$ is the Rician factor determining the strength of the LoS component as compared to the NLoS component. We denote as $\mathbf{H}_t \sim p(\mathbf{H}|\mathbf{D})$ the overall CSI matrix which has the $(j,k)$-th element given by (\ref{Rician fading}), i.e., $[\mathbf{H}_t]_{j,k}=h^{j,k}_t$, with  $p(\mathbf{H}|\mathbf{D})$ being the conditional distribution over the CSI matrix given the distance matrix $\mathbf{D}$.

\subsubsection{Data Generation}\label{data generation}
At each time step $t$, the system collects $N^{\text{PT}}$ CSI matrices\\ $\{\mathbf{H}_{n,t}\}_{n=1}^{N^{\text{PT}}} \mathop{\sim}\limits^{\text{i.i.d.}} p(\mathbf{H}|\mathbf{D}_t)$ given the current distance matrix $\mathbf{D}_t \sim p(\mathbf{D})$. This corresponds to the data points $\{x_{n,t}\}_{n=1}^{N^{\text{PT}}}$ described in the text. Having received information about the context $\mathbf{D}_t$, the DT generates data for context $\mathbf{D}_t$ at a fidelity level dictated by the degree to which the topology modelled by the simulator. Specifically, given a fidelity level $f\in(0, 1)$, the DT simulates a topology characterized by a mismatched distance matrix containing a fraction $f$ of the links as compared to the original distance matrix $\mathbf{D}$. To model this real-to-sim discrepancy, each link of the $K(K-1)$ links is dropped independently with probability $1-f$. The resulting simulation cost $S_f$ is quantified as
\begin{equation} \label{eq:S_f_simulation}
    S_f=K(K-1)fS,
\end{equation}
where $S$ denotes the cost of simulating one edge of the network, and $K(K-1)f$ is the number of edges. 

Given the distance matrix $\mathbf{D}_t$ and the fidelity level $f$, the DT generates CSI matrices $\tilde{\mathbf{H}}_t\in \mathbb{C}^{K\times K} \sim p(\tilde{\mathbf{H}}|\mathbf{D}_t,f)$, where the distribution $p(\tilde{\mathbf{H}}|\mathbf{D}_t,f)$ of the simulated channel (\ref{Rician fading}) depends on the mismatched distance matrix $\tilde{\mathbf{D}}_t$, i.e., $p(\tilde{\mathbf{H}}|\mathbf{D}_t,f) = p(\tilde{\mathbf{H}}|\tilde{\mathbf{D}}_t)$. Accordingly, the synthetic channel matrix $\tilde{\mathbf{H}}$ contains non-zero elements only for a fraction $f$  of links. Furthermore, the non-zero channels are generated using model (\ref{Rician fading}) with possibly mismatched parameters as compared to the true channels.

At each time step $t$, with the remaining budget, DT explores $M^{\text{DT}}$ different distance matrices, i.e.,  $\mathbf{D}_{m,t} \overset{\text{i.i.d.}}{\sim}p(\mathbf{D)}$ for $m=1, \dots, M^{\text{DT}}$, each with $K(K-1)f$ links, from which it generates the corresponding CSI matrices $\tilde{\mathbf{H}}_{m,n,t} \overset{\text{i.i.d.}}{\sim} p(\tilde{\mathbf{H}}|\mathbf{D}_{m,t},f)$ for $n=1,\dots, N^{\text{DT}}$.

\begin{table}[htp]
    \renewcommand\arraystretch{1.4}
	\small
	\caption{Correspondence between Variables in Main Text and Experiment}
	\label{table1}
	\centering
	\scalebox{1}{
	\begin{tabular}{c|c}
		\toprule
		\textbf{Variable in Main Text} & \textbf{Variable in Experiment}\\
		\hline\hline
		\makecell{Context variable: $c_t$} & Distance matrix $\mathbf{D}_t$\\
		\hline
        \makecell{Real-world data from $c_t$: $x_{n,t}$} & CSI matrix $\mathbf{H}_{n,t}$\\
        \hline
		\makecell{Simulated data from $c_t$: $\tilde{x}_{n,t}$} & CSI matrix $\tilde{\mathbf{H}}_{n,t}$\\
		\hline
        \makecell{Simulated context variables: $c_{m,t}$} & Distance matrix $\mathbf{D}_{m,t}$\\
        \hline
        \makecell{Simulated data from  $c_{m,t}$: $\tilde{x}_{m,n,t}$} & CSI  matrix $\tilde{\mathbf{H}}_{m,n,t}$\\
		\bottomrule
	\end{tabular}}
\end{table}
In summary, the corresponding relationship between the variables in the main text and the simulation under the study is summarized in Table \ref{table1}. Specific system parameters in our simulation scenario and the data generation parameters are summarized in Table \ref{table2} and Table \ref{table3}, respectively.
\begin{table}[htp]
    \renewcommand\arraystretch{1.4}
	\small
	\caption{System Parameters}
	\label{table2}
	\centering
	\scalebox{0.85}{
	\begin{tabular}{c|c}
		\toprule
		\textbf{Parameter} & \textbf{Value}\\
		\hline\hline
		Number of transceiver pairs $K$ & $4$\\
		\hline
		Deployment side length $A$ & $100$ m\\
		\hline
		Transceiver pairs distance $d_{\rm{min}}$ and $d_{\rm{max}}$ & $20$ m and $65$ m\\
        \hline
        Path-loss $L(d^{j,k}_t)$ &$148.1+37.6 {\rm{log}}_2(d^{j,k}_t)$\\
        \hline
        Standard deviation of log-normal shadowing $\sigma_s$ & $8$ dB\\
        \hline
        Transmit antenna gain $\eta$ & $9$ dBi\\
        \hline
        \makecell{Rician factor $\kappa$} & $0$ or $5$\\
		\hline
		Noise power $\sigma^2$& $30$ dBm\\
		\hline
		Maximum transmission power $P_{\rm{max}}$& $1$ W\\
		\bottomrule
	\end{tabular}}
\end{table}
\begin{table}[htpb]
    \renewcommand\arraystretch{1.4}
	\small
	\caption{Data Generation Parameters at Time Step $t$}
	\label{table3}
	\centering
	\scalebox{1}{
	\begin{tabular}{c|c}
		\toprule
		\textbf{Parameter} & \textbf{Value}\\
		\hline\hline
		\makecell{Number of CSI matrices at PT $N^{\text{PT}}$} & $10$\\
        \hline
		{Fidelity level at DT $f$} & \makecell{$0.4$}\\
		\hline
        \makecell{Number of simulated CSI matrices at DT $N^{\text{DT}}$} & $20$\\
        \hline
        \makecell{Number of context variables at DT $M^{\text{DT}}$} & $24$\\
		\bottomrule
	\end{tabular}}
\end{table}

\subsection{AMS Mapping and Graph Neural Network-Based Power Control}
Given a context provided by the distance matrix $\mathbf{D}_t$, the AMS mapping aims at determining an AI app that can maximize the sum-rate of all $K$ users by controlling the transmission powers as a function of the CSI. To this end, 
the AMS function maps distance matrix $\mathbf{D}$ to the parameters $\phi$ of the power-controlling GNN. The GNN follows the wireless communication graph convolution network (WCGCN) architecture in \cite{GNN2}. The GNN takes as input the $K\times K$ CSI matrix $\mathbf{H}$ to output a $K\times 1$ transmit power vector $P$. The loss function in (\ref{loss definition}) is specified as the negative sum-rate as in \cite{GNN1,GNN2,GNN3}, i.e., 
\begin{align}
    \ell(\mathbf{H}|\phi) = -\sum_{k=1}^K{\rm{log}}_2\left(1+\frac{|{h_{k,k}}P_k|^2}{\sum_{j\neq k}^K|{h_{j,k}}P_k|^2+\sigma^2}\right),
\end{align}
where $h_{j,k}$ is the $(j,k)$-th entry of matrix $\mathbf{H}$ and $P_k\geq0$ is the $k$-th entry of vector $P$. Further details on the WCGCN employed in this work can be found in Appendix \ref{WCGCN}.  

For the AMS mapping $\phi=g(\cdot|\theta)$, we adopt a multi-layer perceptron (MLP) with two hidden layers equipped with $32$ hidden neurons and exponential linear unit (ELU) as the activation function. The AMS function takes as input a vector of size $K^2\times 1$, obtained as the vectorization of distance matrix $\mathbf{D}$, to output a vector of size $N_\text{GNN} \times 1$, where $N_\text{GNN}$ is the total number of learnable parameters in the WCGCN. For the last layer, we adopt the split-head architecture of reference \cite{hypernetwork} to reduce the computational complexity of the AMS mapping. 

In Algorithm $\ref{A1}$, we set $\lambda_0 = 1$ and $\mu_0 = 0.5$, and the window size $W$ is decreased exponentially over time $t$ from $40$ to $5$, with the window size decreased every time the performance does not improve over two consecutive time steps. The choice of initial hyperparameters $\lambda_0 \textgreater \mu_0$ imposes a higher reliance to the low-variance DT-estimated loss (\ref{DT loss tilde ppi}) at the initial training stage, which we found empirically to be useful for optimization stability, in a manner similar to the observation reported in \cite{initial choice}. The learning rate in (\ref{onlyPT}), (\ref{naive}), (\ref{ppi}), and (\ref{proposed}) are all set as $0.015$. We refer to Appendix \ref{training} for details on training.

\section{Simulation Results}\label{simulation}
In this section, we compare the performances of the DT-powered AMS schemes introduced in Section \ref{DT AMS} and Section \ref{A DT AMS}, alongside the AMS strategy that utilizes only PT data discussed in Section \ref{PT AMS}. To assess the benefits of AMS, we introduce an additional baseline algorithm that does not utilize an AMS mapping, directly optimizing the GNN parameters $\phi$ based on data collected from the physical system for the current context $c$. This baseline uses conventional continual learning (CL), whereby the GNN weights obtained at the previous time $t-1$ are used as the initialization at time $t$ (see, e.g., \cite[Eq. (13.20)]{ML for Engineers}) as
\begin{equation}\label{CL}
    \begin{aligned}
    \phi_t &= \phi_{t-1}- \gamma_{t-1}\nabla\hat{L}^{\text{PT}}(\phi_{t-1}),\\
    \text{with}\  \hat{L}^{\text{PT}}(\phi_{t-1}) &= \frac{1}{N^{{\text{PT}}}}\sum_{n=1}^{N^{{\text{PT}}}}\ell(x_{n,t}|\phi_{t-1}).
\end{aligned}
\end{equation}
Finally, to set a performance upper bound, we consider the weighted minimum mean-square error (WMMSE) power control algorithm \cite{WMMSE}, a benchmark model-based power control scheme that is optimized separately for each CSI matrix $\mathbf{H}$. For each scheme, we report the mean performance and $95\%$ confidence intervals obtained by averaging the results over $5$ random simulation seeds. Simulation code is available at \href{https://github.com/qiushuo0913/DT-powered-AMS}{https://github.com/qiushuo0913/DT-powered-AMS}.

\subsection{Convergence of the Calibration Phase}
In this section, we evaluate the sum-rate achieved under the current context $c_t$ during the calibration phase. To this end, for the CL benchmark, we use the model $\phi_t$ obtained after the update (\ref{CL}) by using data for the current task. In contrast, for the AMS schemes, the sum-rate is evaluated for the GNN model $\phi_t$ obtained by plugging in the current context $c_t$ into the current AMS model as $\phi_t = g(c_t|\theta_t)$. The sum-rate is normalized by the sum-rate obtained under the equal, maximum, power $P_{\rm{max}}$ allocation strategy.

\begin{figure}[bpth]
    \centering
    \includegraphics[width=8cm]{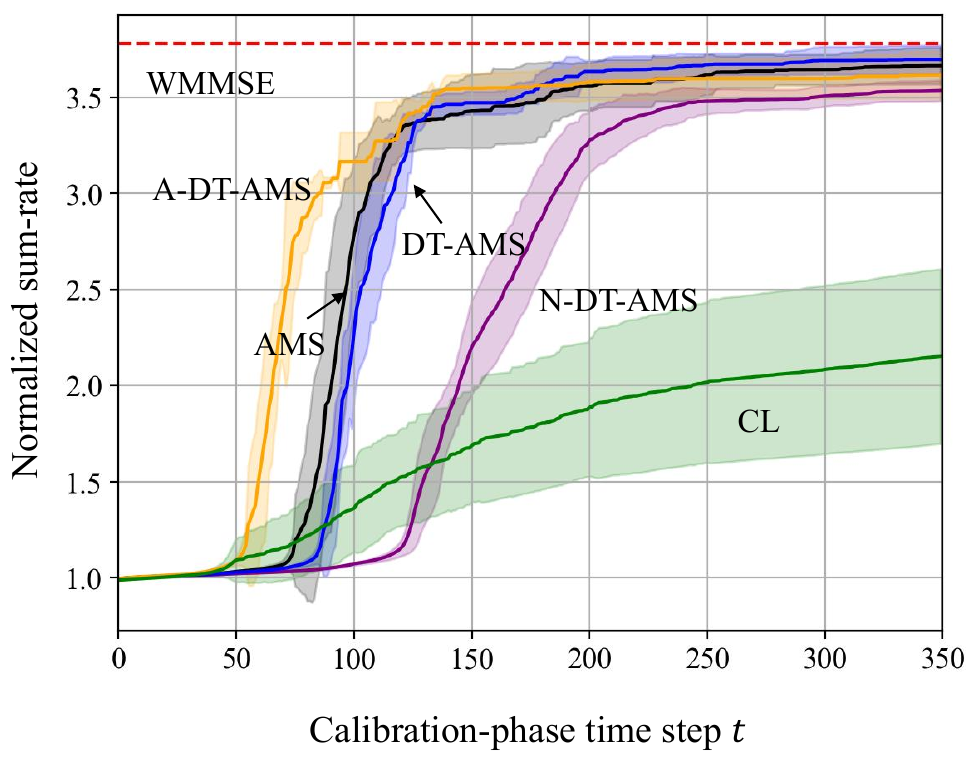}
    \caption{Normalized sum-rate versus the time step $t$ of the calibration phase, with DT channel model matched to the real channel model and real-to-sim discrepancy caused by the fidelity level $f=0.4$.}
    \label{fig:K_facotor0}
\end{figure}
\subsubsection{Digital Twin Model with Well-Specified Channel Model}
We first consider a scenario in which the DT simulator uses the same channel model as the real channel. In this case, the mismatch between simulation and reality is caused by the limited simulation budget that constrains the fraction of simulated links to $f=0.4$. The Rician $\kappa$-factor is set to $\kappa=0$. In Fig. \ref{fig:K_facotor0}, we report the evolution along the time step $t$ of the calibration phase of the normalized sum-rate.

We first note that the conventional continual learning approach converges slowly, failing to reach satisfactory performance. This demonstrates the necessity of employing an AMS strategy that can extract information from the context $c$. In this regard, all AMS schemes are seen to approach the performance of WMMSE for a large enough calibration time step $t$.

That said, the N-DT-AMS scheme has the slowest convergence rate, since it treats equally the simulated and real-world samples. In contrast, the DT-AMS and A-DT-AMS schemes are able to converge faster to the WMMSE normalized sum-rate thanks to the DT-to-PT bias correction term in their loss estimate. In particular, the A-DT-AMS strategy has the best performance, demonstrating the importance of controlling the variance of estimated loss at the early stage of the calibrating phase.

\subsubsection{Digital Twin Model with Misspecified Channel Model} 
\begin{figure}[bpth]
    \centering
    \includegraphics[width=8cm]{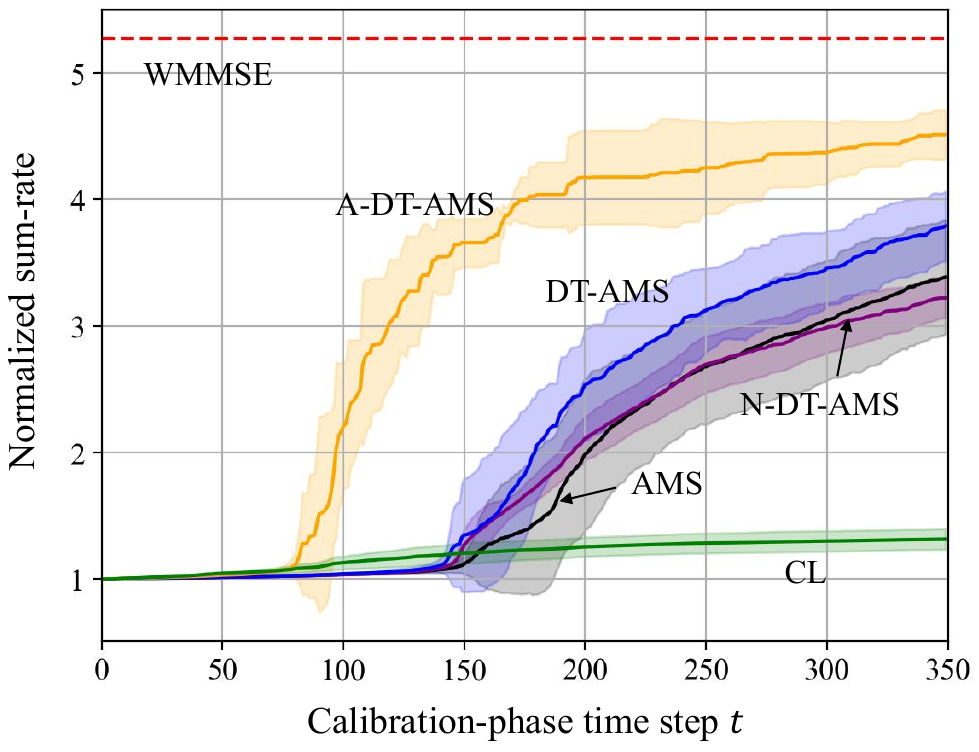}
    \caption{Normalized sum-rate versus the time step $t$ of the calibration phase, with DT channel model using a mismatched Rician $\kappa$-factor in (\ref{Rician fading}) ($f=0.4$).}
    \label{fig:K_facotor5}
\end{figure}

We now consider a misspecified channel model used by the DT. In particular, the DT assumes that the Rician $\kappa$-factor is $\kappa=0$ while the true Rician $\kappa$-factor is $\kappa=5$. As shown in Fig. \ref{fig:K_facotor5}, the proposed A-DT-AMS largely outperforms other AMS strategies and the conventional baseline, converging faster and achieving a higher sum-rate in the early stages of calibration.  This performance gain is justified by the ability of the A-DT-AMS scheme to adaptively control the trade-off between bias and variance as discussed in Section \ref{A DT AMS}. 

\subsection{On the Simulation Cost}
We now investigate the impact of the simulation cost by varying the fidelity level $f$ of the DT simulator in (\ref{eq:S_f_simulation}). In Fig. \ref{fig:different p,B}, we report the normalized sum-rate after a calibration phase of $t=250$ time steps. The performance of DT-powered schemes deteriorates as the fidelity $f$ decreases. However, as compared to the N-DT-AMS scheme, the proposed DT-AMS and A-DT-AMS strategies have a graceful performance degradation thanks to their ability to correct the mismatch between DT and PT data distributions. 

\begin{figure}[bpth]
    \centering
    \includegraphics[width=8cm]{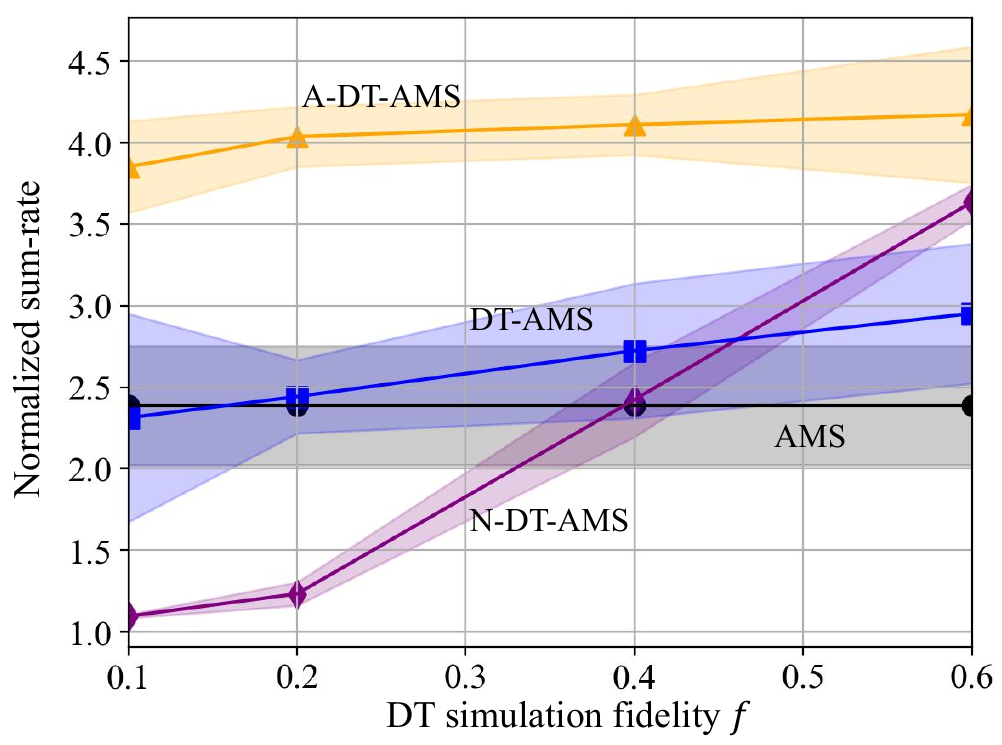}
    \caption{Normalized sum-rate as a function of the DT simulation fidelity $f$ in (\ref{eq:S_f_simulation}).}
    \label{fig:different p,B}
\end{figure}

By (\ref{eq:S_f_simulation}), reducing the fidelity $f$ decreases the simulation cost $S_f$. We now fix the simulation cost $S_f$ to the cost computed with fidelity level $f=0.4$, and adjust the number $M^{\text{DT}}$ of additional contexts simulated by the DT to meet the constraint (\ref{eq:S_f_simulation}). To this end, we set $M^{\text{DT}} =\lfloor 9.6/f \rfloor$. This way reducing the fidelity $f$ permits the simulation of more contexts, defining a trade-off between the diversity of the simulated contexts and the fidelity at which data are generated for each context. Accordingly, unlike in the setting of Fig. \ref{fig:different p,B}, a fidelity reduction does not necessarily imply a performance degradation, as it allows to the DT to generate more contexts under the same simulation budget $S_f$. 

The results shown in Fig. \ref{fig:different p} demonstrate that N-DT-AMS still exhibits a monotonically increasing rate as a function of $f$, since it is unable to compensate for the reduced fidelity of the simulation at DT. In contrast, thanks to the inclusion of DT-to-PT correction terms, both A-DT-AMS and DT-AMS can withstand a reduction in the simulation fidelity $f$, benefiting from access to a larger synthetic data set covering more contexts.

\begin{figure}[bpth]
    \centering
    \includegraphics[width=8cm]{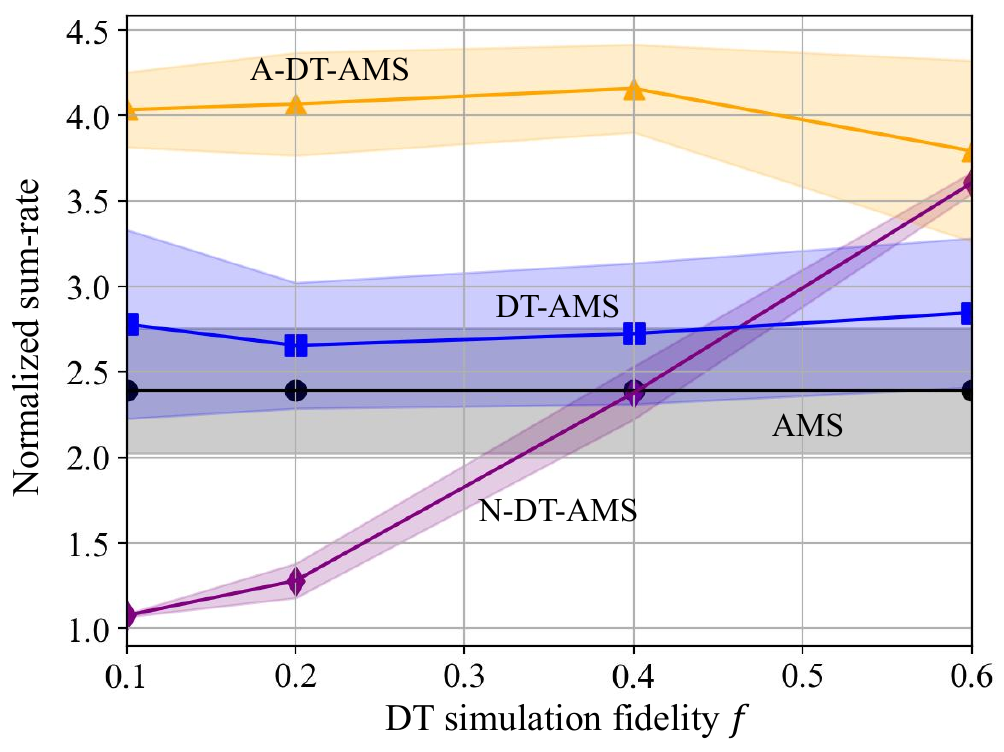}
    \caption{Normalized sum-rate as a function of the DT simulation fidelity level $f$ in (\ref{eq:S_f_simulation}) for a fixed simulation budget.}
    \label{fig:different p}
\end{figure}

\subsection{Impact of the Numbers of Context Variables and Simulated Data per Context Variable}
\begin{figure}[bpth]
    \centering
    \includegraphics[width=8cm]{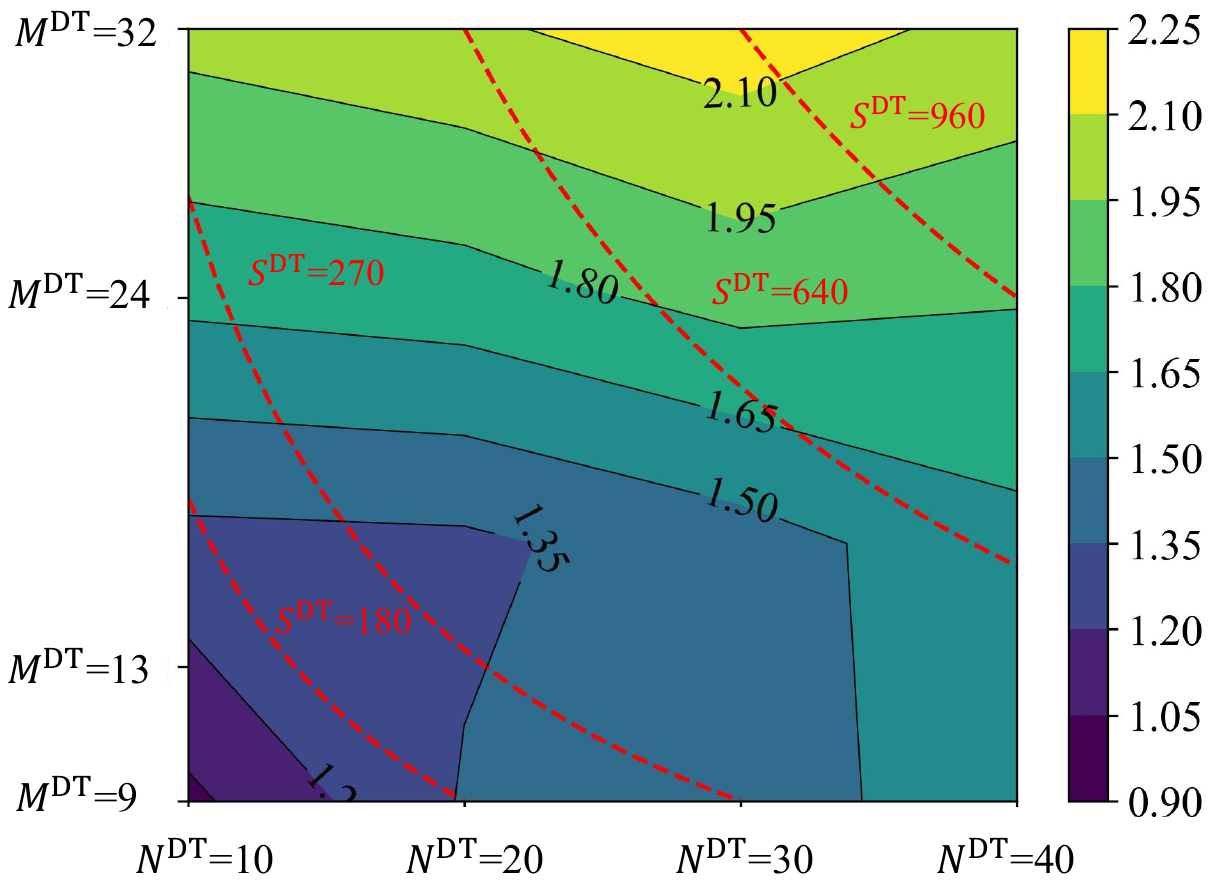}
    \caption{Sum-rate of A-DT-AMS, normalized by the sum-rate of AMS, as a function of the number of simulated contexts, $M^{\text{DT}}$, and of the number of data points per context, $N^{\text{DT}}$. }
    \label{fig:heatmap}
\end{figure}

Given a fixed simulation budget (\ref{eq:S_f_simulation}) and a fixed fidelity level $f$, one can explore different choices for the number of generated contexts, $M^{\text{DT}}$, and for the number of simulated data points per context, $N^{\text{DT}}$. For fidelity $f=0.4$, Fig. \ref{fig:heatmap} reports the sum-rate of A-DT-AMS after $t=100$ calibrating steps. This figure normalizes the sum-rate by the value achieved by the baseline AMS scheme, which does not use DT data. Red dashed lines correspond to different budget levels $S^{\text{DT}}$, namely $S^{\text{DT}}=180$, $S^{\text{DT}}=270$, $S^{\text{DT}}=640$, and $S^{\text{DT}}=960$. 

It is observed from Fig. \ref{fig:heatmap} that, an increased simulation budget $S^\text{DT}$ generally increases the performance of A-DT-AMS by virtue of the reduced variance of the DT-related loss terms used for online training. Furthermore, it is generally beneficial to increase the number of contexts $M^{\text{DT}}$ rather than the number of CSI realizations $N^{\text{DT}}$. This suggests that a large diversity of contexts is preferable to more data per context.

\subsection{On the Correlation of Context Variables}\label{markov_simulation}
Finally, we study the impact of memory in the context variables $c_t$ for $t=1,2,\dots$ To this end, we consider a setting in which, at time step $t=0$, Txs and Rxs are placed as in the previous subsection. However, in the following time steps $t=1,2,\dots$, we keep each Tx and Rx fixed with some probability $p$, while with probability $1-p$ the Rx moves in a random uniformly sampled position around the Tx within a circle with a radius equal to the current distance between the Rx and the Tx. This situation is meant to capture the downlink of cellular deployments in which mobile devices can move within the coverage area of the serving base station.

At the DT, we consider two simulation strategies. The first, adopted in the previous subsections, draws i.i.d. samples from the stationary context distribution, thus generating synthetic Tx and Rx positions from the true deployment distribution at time $t=0$. The second approach starts from the current context $c_t$ to generate contexts from the Markov mobility model described above. This way, with the second strategy, the DT leverages knowledge about the temporal correlation of the context variables.

\begin{figure}[htp]
    \centering
    \subfigure[]{
    \includegraphics[width=7.5cm]{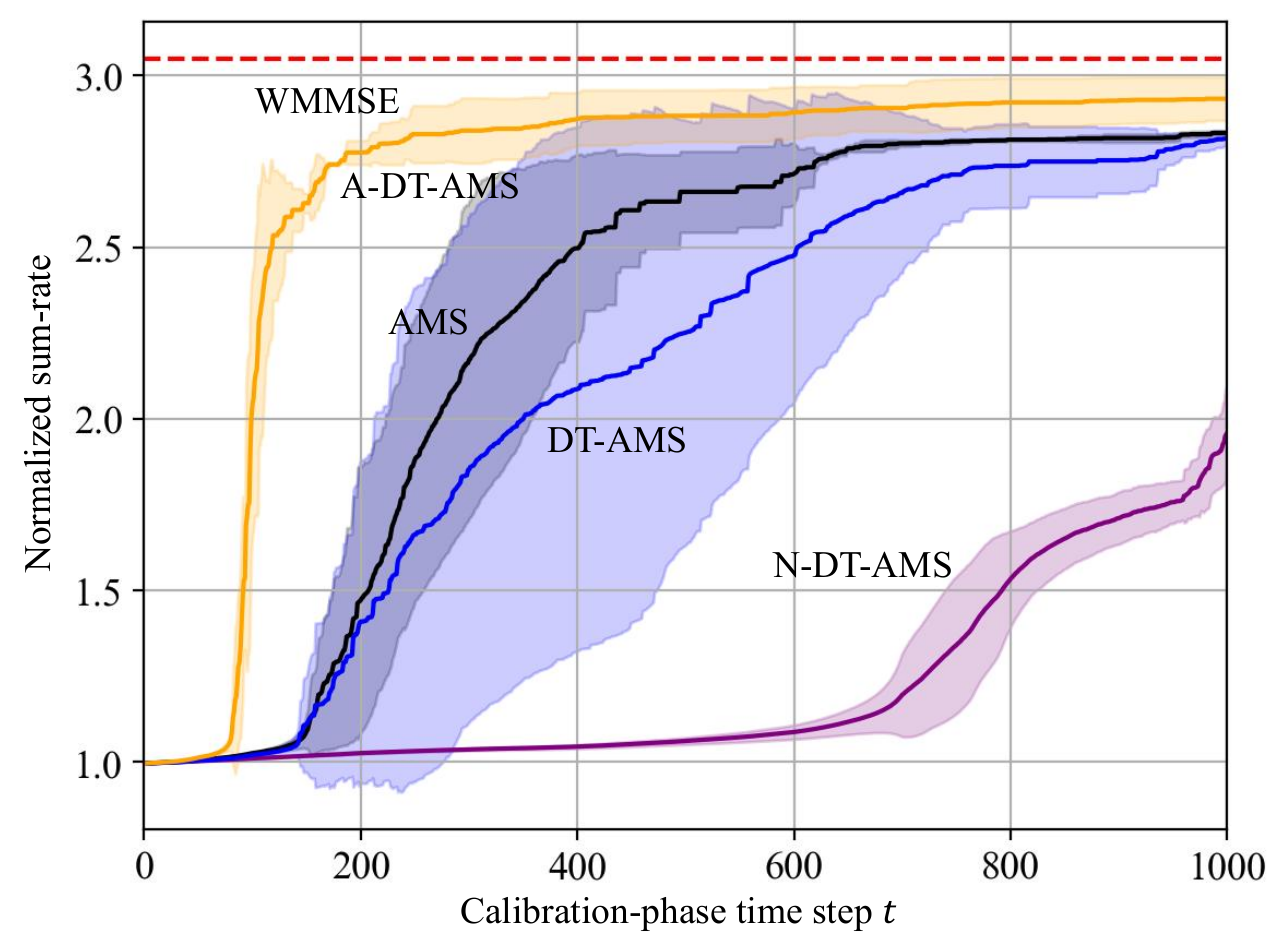}
    \vspace{-0.5cm}
    \label{fig:DT_PT_markov}
    }
    \subfigure[]
    {
    \centering
    \includegraphics[width=7.5cm]{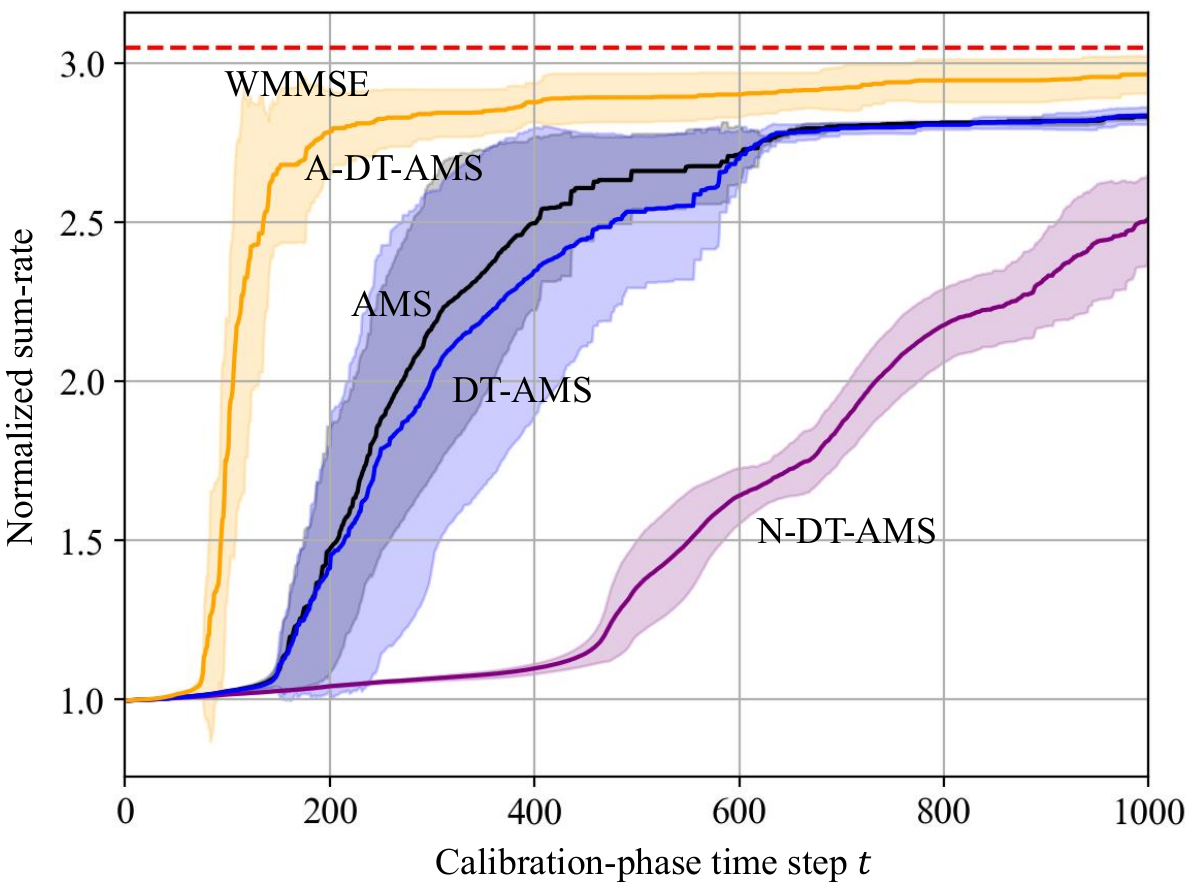}
    \label{fig:DT_i.i.d._PT_markov}
    }
    \caption{Normalized sum-rate versus the time step $t$ of the calibration phase, with Markovian context variables. The left figure shows the performance of DT-based schemes that simulate context variables in an i.i.d. fashion, while the right figure considers a DT that can generate contexts via the Markovian mobility model followed by the true context process.}
    \label{fig: markov}
    \vspace{-0.5cm}
\end{figure}

Fig. \ref{fig: markov} presents the convergence performance for both DT sampling strategies, namely i.i.d. (left) or Markov (right). The results confirm the main conclusions presented in the previous subsection, showing that the proposed A-DT-AMS approach has the best performance, even under context processes with memory. Moreover, comparing the left and right figures, all DT-based approaches that leverage knowledge of the temporal correlation in the context process, shown on the right, have a faster convergence than methods that assume i.i.d. context variables, presented on the left.

\section{Conclusion}\label{conclusion}
Next-generation wireless systems are expected to make extensive use of AI-based apps. The deployment of AI-based apps is highly context dependent, with different network conditions, user requirements, and operator's intents requiring an adaptation of the apps. In order to ensure the adaptation to the current context, an automatic model selection (AMS) function maps context variables to model parameters. Training an effective AMS generally requires exposure to a sufficiently large amount of contexts, calling for a potentially long online calibration phase. This work has proposed a principled way of speeding up the calibration of an AMS model by utilizing data generated from a digital twin (DT) from multiple synthetic contexts. The proposed calibration schemes, referred to as DT-AMS and A-DT-AMS, move beyond conventional solutions that use synthetic data to augment the data set. This is done by utilizing real-world data to correct the bias caused by the discrepancy between the DT simulator and the real world. Simulation results have demonstrated the advantages of the proposed DT-powered AMS calibration approaches. 

Future work may focus on settings in which multiple apps must be selected, managing conflicts between apps; on reducing the communication overhead between PT and DT; on applying transformers as in-context learners as AMS mappings \cite{in-context}; as well as on updating context distribution at the DT side via continual feedback from PT \cite{conclusion Data shift}.

\appendices

\section{Proof of the Solution of (\ref{lambda mu problem})} \label{lambda mu solution}
We propose to tune the hyperparameters $\lambda_t$ and $\mu_t$ so as to minimize the MSE, i.e., 
\begin{equation}\label{MSE_appendix}
    \begin{aligned}
    &\{\lambda_t^{*}, \mu_t^{*}\} = \mathop{\rm{argmin}}\limits_{\{\lambda_t,\mu_t\}}\ \mathrm{MSE}(\theta_t),\\
    &=  \mathop{\rm{argmin}}\limits_{\{\lambda_t,\mu_t\}}\ 
    \begin{bmatrix}
    \lambda_t&\mu_t
    \end{bmatrix}
    \\&\underbrace{\begin{bmatrix}
        {L^{\text{DT}}(\theta_t)}^2+V^{\text{DT}}(\theta_t) & -{L^{\text{DT}}(\theta_t)}^2 \\
        -{L^{\text{DT}}(\theta_t)}^2 & {L^{\text{DT}}(\theta_t)}^2+V^{\text{PT}\rightarrow\text{DT}}(\theta_t)
    \end{bmatrix}}_{Q}
    \begin{bmatrix}
    \lambda_t\\
    \mu_t
    \end{bmatrix}\\ &+ 
    \underbrace{\begin{bmatrix}
        0&-2C^{\text{PT}\rightarrow\text{DT}}(\theta_t)
    \end{bmatrix}}_{b}
    \begin{bmatrix}
    \lambda_t\\
    \mu_t
    \end{bmatrix}
    +V^{\text{PT}}(\theta_t).
\end{aligned}
\end{equation}

Note that $Q$ is always positive definite, thus, the aforementioned optimization problem is strictly convex on $\lambda_t$ and $\mu_t$. Then, we can obtain the unique optimal solutions for $\lambda_t$ and $\mu_t$ based on the formula
\begin{equation}
   \begin{bmatrix}
    \lambda_t\\
    \mu_t
    \end{bmatrix}
    =
    -\frac{1}{2}Q^{-1}b,
\end{equation}
i.e.,
\begin{equation}
        \begin{aligned}
            \lambda_t^* &= \frac{C^{\text{DT}\rightarrow\text{PT}}(\theta_t){L^{\text{DT}}(\theta_t)}^2}{{L^{\text{DT}}(\theta_t)}^2\left(V^{\text{DT}\rightarrow \text{PT}}(\theta_t)+V^{\text{DT}}(\theta_t)\right)+V^{\text{DT}\rightarrow \text{PT}}(\theta_t)V^{\text{DT}}(\theta_t)},\\
            \mu_t^* &= \frac{C^{\text{DT}\rightarrow\text{PT}}(\theta_t)+\lambda_t^*{L^{\text{DT}}(\theta_t)}^2}{V^{\text{DT}\rightarrow \text{PT}}(\theta_t)+{L^{\text{DT}}(\theta_t)}^2}.
        \end{aligned}
    \end{equation}

Assume now that the context distribution used by the DT for sampling, which is denoted as $\hat{p}(c)$, is different from the true context distribution $p(c)$. Note that distribution $\hat{p}(c)$ represents the stationary distribution adopted by the DT to sample context variables. Denote as $\epsilon = D_{\mathrm{TV}}(\hat{p}(c), p(c))$ the total variation (TV) distance between $\hat{p}(c)$ and $p(c)$. With this definition, the inequality $\mathbb{E}_{\hat{p}(c)}[z] \leq \mathbb{E}_{p(c)}[z]+z_{\rm{max}}$ holds for any random variable $z$ taking maximal value $z_{\rm{max}}$\cite{ML for Engineers}. Using this inequality to evaluate an upper bound on the MSE in (\ref{MSE_appendix}), one can follow the same steps above to obtain optimized values
    \begin{equation}\label{lambda mu revise}
        \begin{aligned}
            \lambda_t^* &= \frac{C^{\text{DT}\rightarrow\text{PT}}(\theta_t)({L^{\text{DT}}(\theta_t)}^2+\epsilon L_{\rm{max}}L^{\text{DT}}(\theta_t))}{{L^{\text{DT}}(\theta_t)}^2\left(V^{\text{DT}\rightarrow \text{PT}}(\theta_t)+V^{\text{DT}}(\theta_t)\right)+V^{\text{DT}\rightarrow \text{PT}}(\theta_t)(\epsilon^2{L^2_{\rm{max}}}+2\epsilon L_{\rm{max}}L^{\text{DT}}(\theta_t))+V^{\text{DT}\rightarrow \text{PT}}(\theta_t)V^{\text{DT}}(\theta_t)},\\
            \mu_t^* &= \frac{C^{\text{DT}\rightarrow\text{PT}}(\theta_t)+\lambda_t^*{(L^{\text{DT}}(\theta_t)}^2+\epsilon L_{\rm{max}}L^{\text{DT}}(\theta_t))}{V^{\text{DT}\rightarrow \text{PT}}(\theta_t)+{L^{\text{DT}}(\theta_t)}^2},
        \end{aligned}
    \end{equation}
where $L_{\rm{max}}$ is the maximum value of the loss function. Note that the original values $\lambda^*_t$ and $\mu^*_t$ in (40) are recovered when $\epsilon = 0$.

\section{Implementations Details on WCGCN} \label{WCGCN}
The WCGCN model \cite{GNN2} belongs to the class of message passing graph neural networks. WCGCN treats its input CSI matrix $\mathbf{H}_t$ via the corresponding graph representation $\mathcal{G}$. Specifically, the graph corresponds to the tuple $\mathcal{G} = (\mathcal{V}, \mathcal{E}, \Zstroke, \mathbf{A})$, where $\mathcal{V}$ is the set of nodes, $\mathcal{E}$ is the set of edges, $\mathbf{A}$ is the adjacency feature matrix, and $\Zstroke$ is a feature vector, including the features of all the nodes. Following the modeling in \cite{GNN2}, each node $k \in \mathcal{V}$ in the graph corresponds to the $k$-th transceiver pair $(T^k_t,R^k_t)$; each edge $(j,k) \in \mathcal{E}$ represents the interference from node $j$ to node $k$, i.e., $(j,k) \in \mathcal{E}$ if $h_t^{j,k} \neq 0$ and $(j,k) \notin \mathcal{E}$ if $h_t^{j,k} = 0$ ; each  node feature vector $\Zstroke\in \mathbb{C}^{K}$ corresponds to the diagonal elements of $\mathbf{H}_t$, i.e., $[\Zstroke]_k = h^{kk}_t$; and finally the adjacency feature matrix $\bm{A}\in \mathbb{C}^{K \times K}$ is defined based on the off-diagonal terms of $\mathbf{H}^t$, i.e., $[\mathbf{A}]_{j,k} = h_t^{jk}$ for $j \neq k$. 

WCGCN updates its hidden state $\mathbf{u}_k^i$ via two multi-layer perceptrons (MLPs) by taking into account for the two important operations in message passing, i.e., aggregation and combining, which can be expressed as
\begin{small}
\begin{equation}
 \begin{aligned}
    \mathbf{u}^i_k &= \beta\left({\rm{MLP_2}}\left(\mathbf{u}_k^{i-1}, {\rm{MAX}}_{j\in\mathcal{N}(k)}\left\{{\rm{MLP_1}}\left([\mathbf{u}_j^{i-1}, [\mathbf{A}]_{j,k}]\right)\right\}\right)\right),\\  &1\leq i \leq I. \label{eq:WCGCN_detail}
\end{aligned}
\end{equation}
\end{small}
In (\ref{eq:WCGCN_detail}), ${\rm{MLP_1}}$ and ${\rm{MLP_2}}$ represent the two different MLPs, respectively;  $\mathbf{u}_k^{i}$ denotes the hidden state of node $k$ at the $i$-th layer of WCGCN; $\beta(\cdot)$ is a differentiable normalization function to constrain the power $P_k$; $\mathcal{N}(k)$ denotes the set of the neighbors of node $k$; and $I$ denotes the number of layers in WCGCN.

\section{Training Details} \label{training}
All experiments are implemented in Python 3.9 with Pytorch 1.12.1, where the GNN is implemented by Pytorch Geometric 2.5.0. For fair comparison, we use the same training policy for all approaches. Specifically, we use the SGD optimizer with a weight decay factor of $0.01$ and decrease the learning rate by dividing its value by a factor of $2$ for every $50$ epochs during the online training while setting the initial learning rate to $0.015$. The batch sizes used for a single SGD update are the same as the values of $N^{\text{PT}} = 10$ and $N^{\text{DT}} = 20$ for PT and DT, respectively.

\end{document}